\documentclass[10pt]{article}

\usepackage{mathrsfs} 

\usepackage[]{graphicx}    
\usepackage[gen]{eurosym}
\usepackage{verbatim}
\usepackage{amsthm}
\usepackage{amsmath}
\usepackage{dsfont}
\usepackage{amssymb}
\usepackage{latexsym}		
\usepackage{tabularx}
\usepackage{setspace}
\usepackage{listings}
\usepackage{stmaryrd}
\usepackage{hhline}
\usepackage{booktabs}
\usepackage{caption}

\usepackage{arxiv}

\usepackage[utf8]{inputenc} 
\usepackage[T1]{fontenc}    
\usepackage{hyperref}       
\usepackage{url}            
\usepackage{booktabs}       
\usepackage{amsfonts}       
\usepackage{nicefrac}       
\usepackage{microtype}      
\usepackage{natbib}

\usepackage[capposition=top]{floatrow}
\usepackage{multirow,lscape,longtable}

\newcommand{\cay}[1]{[\citeauthor{#1} \citeyear{#1}]}
\newcommand{\cayNP}[1]{\citeauthor{#1} (\citeyear{#1})}

\title{A SURVEY ON NATURAL LANGUAGE PROCESSING (NLP) \& APPLICATIONS IN INSURANCE}

\author{

 Antoine Ly \\
  Data Analytics Solutions\\
  SCOR \\
  Paris, France \\
  \texttt{aly@scor.com} \\
 \And
 Benno Uthayasooriyar \\
  Université de Bretagne Occidentale\\
  EURIA\\
  Brest, France \\
  \texttt{uth.benno@gmail.com} \\
  \And
 Tingting Wang \\
  Data Analytics Solutions \\                          
  SCOR\\
  Beijing, China \\
  \texttt{twang@scor.com} \\ 
}

\begin{document}
\maketitle

\begin{abstract}
Text is the most widely used means of communication today. This data is abundant but nevertheless complex to exploit within algorithms. For years, scientists have been trying to implement different techniques that enable computers to replicate some mechanisms of human reading. During the past five years, research disrupted the capacity of the algorithms to unleash the value of text data. It brings today, many opportunities for the insurance industry. 

Understanding those methods and, above all, knowing how to apply them is a major challenge and  key to unleash the value of text data that have been stored for many years. Processing language with computer brings many new opportunities especially in the insurance sector where reports are central in the information used by insurers.

SCOR's Data Analytics team has been working on the implementation of innovative tools or products that enable the use of the latest research on text analysis. Understanding text mining techniques in insurance enhances the monitoring of the underwritten risks and many processes that finally benefit policyholders.

This article proposes to explain opportunities that Natural Language Processing (NLP) are providing to insurance. It details different methods used today in practice\footnote{And especially those implemented in the different tools we propose to our clients} traces back the story of them. We also illustrate the implementation of certain methods using open source libraries and python codes that we have developed to facilitate the use of these techniques.

After giving a general overview on the evolution of text mining during the past few years, we share about how to conduct a full study with text mining and share some examples to serve those models into insurance products or services. Finally, we explained in more details every step that composes a Natural Language Processing study to ensure the reader can have a deep understanding on the implementation.
\end{abstract}


\section*{Introduction}

\section{What is it about?}

Text mining is a field related to data analytic consisting in the analysis of textual data. A textual data point can be a character, a word, a sentence, paragraph or a full document. As text data is unstructured (opposed to tabular data), it requires specific approaches and models to be able to use it. 

Most of the time, text is analysed by a human who can read the text and transform it into structured information. It can take a lot of time when it comes to analyzing thousands of sentences. Natural Language Processing (NLP) refers to the capacity of copying the human reading process. Both terms "Text mining" or "Natural Language Processing" are used in the industry to refer to text manipulation with algorithms (despite some differences between both concepts). More than just mining text, NLP also tries to capture the complexity of the language and translates it into summarized information that a computer could use. 

\begin{figure}[H]
  \centering
  \includegraphics[scale=0.45]{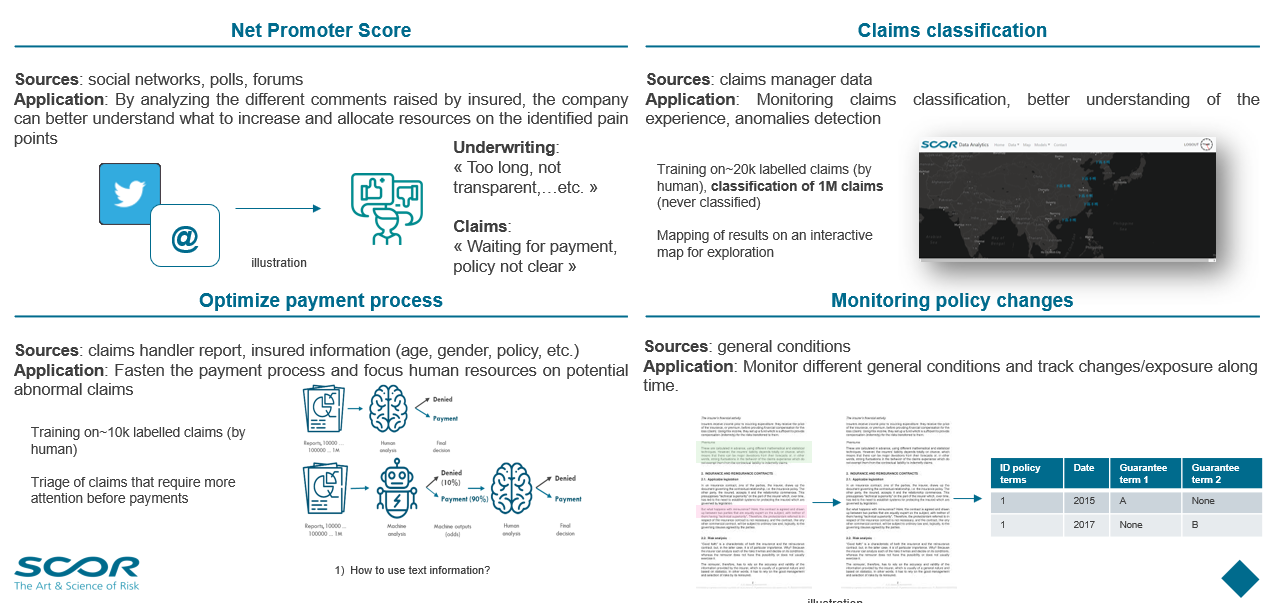}
  \caption{Usage example of Natural Language Processing based on SCOR experience.}
  \label{exampleNLP}
\end{figure}

Nowadays, text is everywhere. It is the main information we all use. Text is present in every website, report, or any other digital mean of communication. Thus, Natural Language Processing paves the way towards numerous opportunities starting to be heavily used by the insurance industry as illustrated on Figure \ref{exampleNLP}.

\section{Opportunities for the insurance industry.}

In insurance, NLP can be leveraged at many steps of a policy life cycle. We present below some examples. Details on some use cases are also presented later in this article.

\begin{itemize}
    \item \textbf{Marketing}: NLP has been used to enhance marketing actions for a couple of years by different insurance companies. At SCOR, we used it to monitor the sentiment analysis of comments to better consider insured needs or monitor the perception on a specific risk (e.g. what people are thinking about a particular product). It is common also to process feedback that people publish on different social networks such as Twitter, Reddit or press articles to extract some trends in expectations and then better pilot the strategy of the company. Automating comments analysis enable a live monitoring of what is published on the internet. NLP model can indeed convert text into a general sentiment score or extract the topic they are referring to. If the underlying NLP technique can be sophisticated and have been evolving, it performs pretty well and the challenge is more about integrating such models into operational daily business. We illustrate this challenge and some solutions in paragraphs below.
    \item \textbf{Underwriting}: At the re-insurance level, underwriters need to analyse a large number of policies at each renewal period. After digitalization of the contract, many text mining applications are possible on the wordings in order to track changes and check compliance. It makes the underwriting process friction-less and enable a better monitoring and change tracking in clauses. Those automatic checks might also be useful for the insured to provide more transparency on the coverage of the policies from one period to another. In life insurance, many opportunities are also opened such as enhancing medical report analysis. By coupling Optical Character Recognition (OCR)\footnote{A paper focusing of this topic will be released soon} and NLP we can indeed extract information from Medical reports and help underwriters to focus on most difficult cases while others can take less time to process. In another coming paper, we will detail how for instance NLP can be used to automatically anonymize documents with Personal Information and then process data compliant with GDPR. We explain in this article one of the key NLP method that can be used for anonymization which is called Name Entity Recognition (NER).
    \item \textbf{Claims}: Textual analysis of claims and classification are two topics where SCOR has been using NLP. It can simplify acceptance of claims in the process to reduce time treatment, operational errors or even help in fraud detection. We detail below how text mining can be used for claims adjudication or to optimize human reading to classify claims and route them to the appropriate department. 
    \item \textbf{Reserving}: Analyzing comments of claim reports during the First Notification of Loss or expert claims assessment is more and more developed in insurance company to improve reserving for severe claims. Descriptions can provide valuable information to anticipate the development of claim and better estimate the expected cost. In life insurance, the assessment of expert reports could help in reserve projection. It could for instance help in Long Term Care or specific Critical Illness products where risk factors can be cross correlated to different dependent diseases. Despite the low annual frequency of such reports, the amount of document can become significant on a long time study period for which a systematic approach is useful.
    \item \textbf{Prevention}: Enhancing Medical Diagnostic is another opportunity for NLP. It could help doctors to ensure they cover all possible diseases matching with symptoms. In this particular field, the challenge is to train NLP models on medical data that have a specific vocabulary. Early diagnostic is really key to ensure good health of people and avoid complications. Some systems already exists to capture different fields but they are quite time consuming. Doctors have to fill many fields that are structured in templates and most of the time free comments are not allowed (while they are the most easy to use). In such application, NLP models must be applied in compliance with highly secured systems (anonymization of data, local computation, etc.). We detail further in this article some technologies that can be used to embed NLP model into solutions and help in respecting such constraints.  
\end{itemize}

Manipulating text, provides many new opportunities in the insurance industry. We explain further in this paper how to apply the different techniques of Natural Language Processing in order to deliver business use cases. The article gradually explains the different steps that are commonly used in text manipulation and illustrates how new state-of-the art NLP models can help insurance industry to process for instance claims description, or to adjudicate claims based on report and then optimize payment process (by focusing investigation on most relevant cases). However, if research has been doing tremendous progress in text manipulation, it is not without challenges to apply the different techniques to the industry and provide NLP based product.

\section{Main challenges to embed NLP models into solutions.}

\subsection{How to access or collect useful text data?}

Text data are almost everywhere. We use them to communicate and send information that are stored into different layers: emails, reports, websites, information systems, logs, etc. However, they are also one the less exploited information by industries. If data are stored everywhere, gathering them (i.e. relevant text information) can be a real challenge. In legacy industries like insurance, text data are stored within paper documents that most of the time are scanned. Capturing and extracting those information is the first challenge to be able then to use text into NLP models. Some methods based on Optical Character Recognition (OCR) that SCOR implemented to overcome that first extraction step will be detailed in another paper coming in Q4 2020. \\

For most recent digital systems, text is easier to access and can be either collected directly from databases lying behind applications, websites, social media or by simply converting documents (like pdf generated by an information system). When it comes to collecting data from social media, like Twitter, Reddit, etc. those are usually providing some connectors (APIs) that allow to collect information in a more structured format. In any cases, as for regular tabular data, the collect of text must be driven by an initial business issue. Collecting text information without restriction can indeed lead to bias and noise that algorithms might struggle to deal with since relations between text and business issue will be harder to detect.\\

\subsection{Enrich traditional information to enhance services.}

In practice, text data are mainly used in addition to more traditional tabular data. For example, to speed up payment process, free text that describes the claim can be added to tabular data that contain more traditional information (policy, claim date, type, etc.). Text is usually really rich information that humans process to take decision or to explain context. This specific context is most of the time a piece of information that is difficult to capture within structured data.

\begin{figure}[H]
    \centering
    \includegraphics[width=15cm]{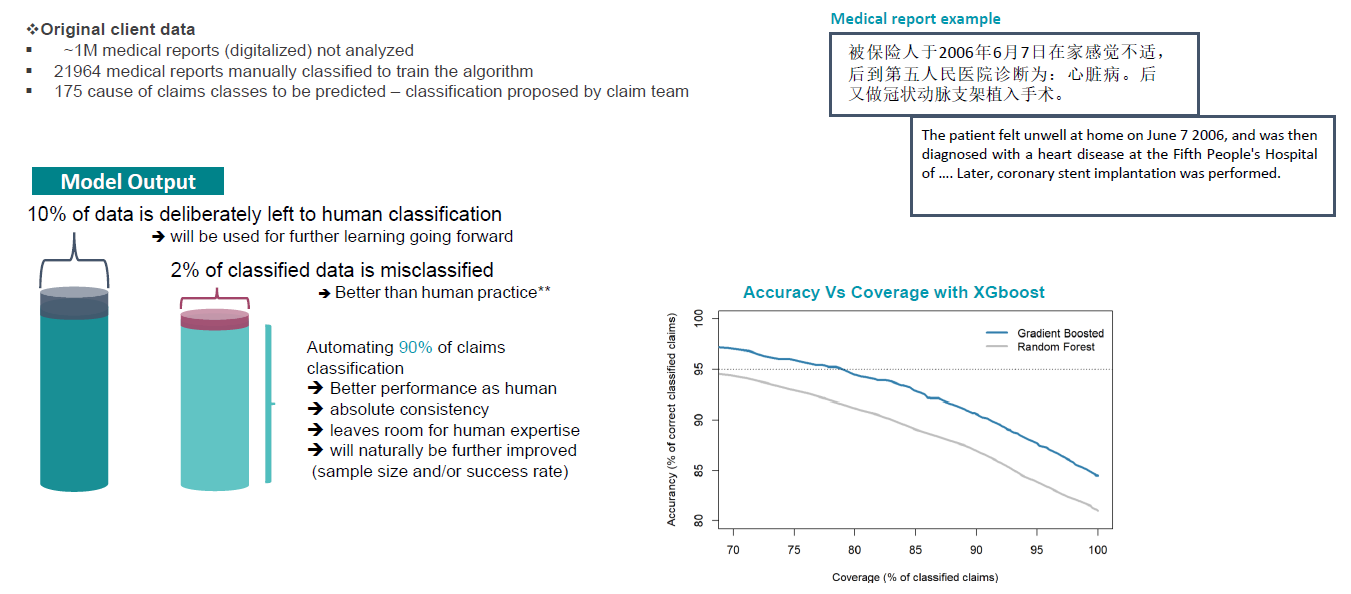}
    \caption{Critical illness classification use case}
    \label{Use_case}
\end{figure}

Since text is unstructured, NLP helps in creating numerical representation of language so it can be merged to standard tabular information. The demand in the insurance market in the usage of text is high. As one use case, we present in the following an application which was developed to showcase the added-value of NLP (deep learning) algorithms in the classification of claims and SCOR capacity to interface such models with insurance practitioner. The following example shares how to automate Critical Illness Claims classification and achieve performance equivalent (if not even better) to human practice. The usage of NLP models increases the efficiency of claim analysis and accelerates the claims classification process. Thus with 20 000 reports, our algorithm classified accurately 90\% of them (cf Figure \ref{Use_case}). \\

The fitted algorithm can then be used to process the full history of claims and to monitor underwriting network distribution and the evolution of claims by location.\\

NLP models integration helps not only the insurance company to get a better insights on their claims but also allows to remove friction in the claim payment process for regular claims. With the apparition of more and more digital insurance during the past few years, increasing  every step of an insurance product life cycle became necessary. NLP is definitely a key asset here to better use free text and enhance the reactivity of insurers.

\subsection{Interfacing NLP models.}

When it comes to use machine learning models, integrating them into real products can be challenging. To overcome the proof of concept, models need to be integrated in product pipelines and be monitored properly. For end users, models need to be used through simple interface that can make decisions easier but also transparent enough to avoid any black box effect.

\begin{figure}[H]
    \centering
    \includegraphics[width=15cm]{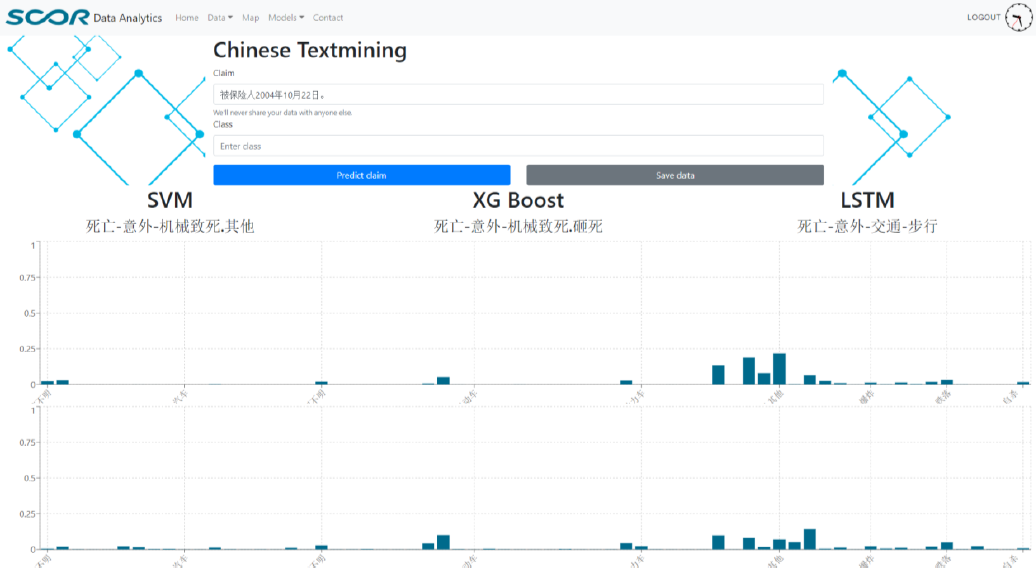}
    \caption{Web application for critical illness classification}
    \label{UI_chinese_textmining}
\end{figure}

Figure \ref{UI_chinese_textmining} illustrates the web interface that has been developed to classify claims. More specifically, the model was integrated in one back-end application and we then developed a tailored full-stack, secure and cloud based solutions for our customer (following state-of-the art practices detailed below).\\

It ensures end users to control the classification and to use the model to label historical data. Such an interface is particularly useful for instance if we want to visualize and monitor specific distribution networks and study if underwriting can be improve in specific locations.

In practice, in order to setup the environment, SCOR Analytics team is using Docker. It makes it easier to create, deploy, and run applications by using containers. Docker is an open source technology that can be seen as a lightweight virtual machine. One Docker container can be used for the machine learning algorithm development: data preprocessing, embedding and modeling, ML pipelines, results analysis. It helps in tracking the environment and reviewing. Once the complete pipeline is in place, the trained model is saved as well as its environment. Thus, for the claim classification use case, we embedded a machine learning model into another container, using the Python library Flask\footnote{\url{https://flask.palletsprojects.com/en/1.1.x/}} to build a Web API (Application Programming Interface). It allows the user to access the pipeline with their own data by making an HTTP request. It's also possible to develop a user-friendly graphical Web application using JavaScript frameworks like Angular\footnote{\url{https://angular.io/}} or React\footnote{\url{https://reactjs.org/}}.

\begin{figure}[H]
    \centering
    \includegraphics[width=15cm]{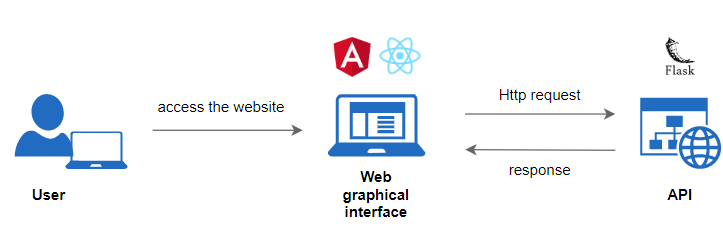}
    \caption{User flow diagram}
    \label{Userflow}
\end{figure}

\begin{figure}[H]
    \centering
    \includegraphics[width=15cm]{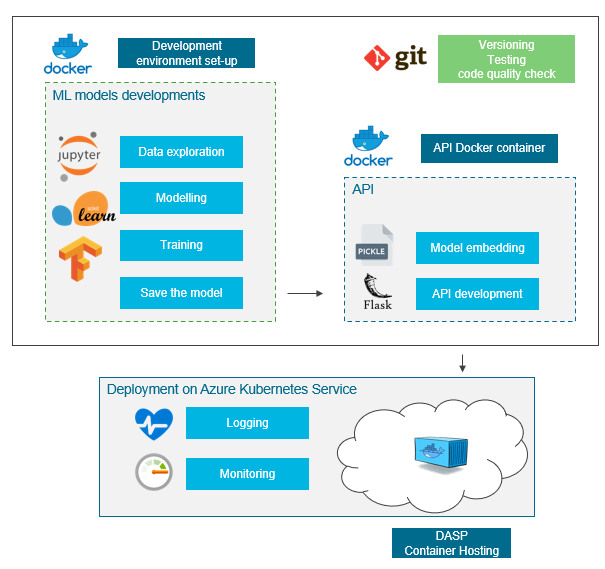}
    \caption{Deployment architecture diagram}
    \label{Deployment_architecture}
\end{figure}

The Figures \ref{Userflow} and \ref{Deployment_architecture} illustrates the logical view of the process and architecture used to ensure modular implementation of the model. This architecture is not specific to text mining models but illustrate more broadly what technologies can be used to deploy smoothly machine learning models into an existing process.

\subsection{Some new opportunities with latest releases in NLP?}

Along the previous paragraphs we mentioned sentiment analysis and claims classification as some examples. SCOR has been using text mining on multiple lines of business for couple of years now. It brings value not only for internal process but also to insured since it speed-up processes. \\

Another example is the claims adjudication. In this use case, text mining is used to strengthen the traditional tabular data. By coupling text embedding to claims profile, we designed an algorithm able to speed-up payments. A score is  used to design 3 priority categories (High, Medium, and Low) and help claim managers to prioritize claims investigation on complex cases and automatize payment for regular claims.\\

\begin{figure}[H]
    \centering
    \includegraphics[width=15cm]{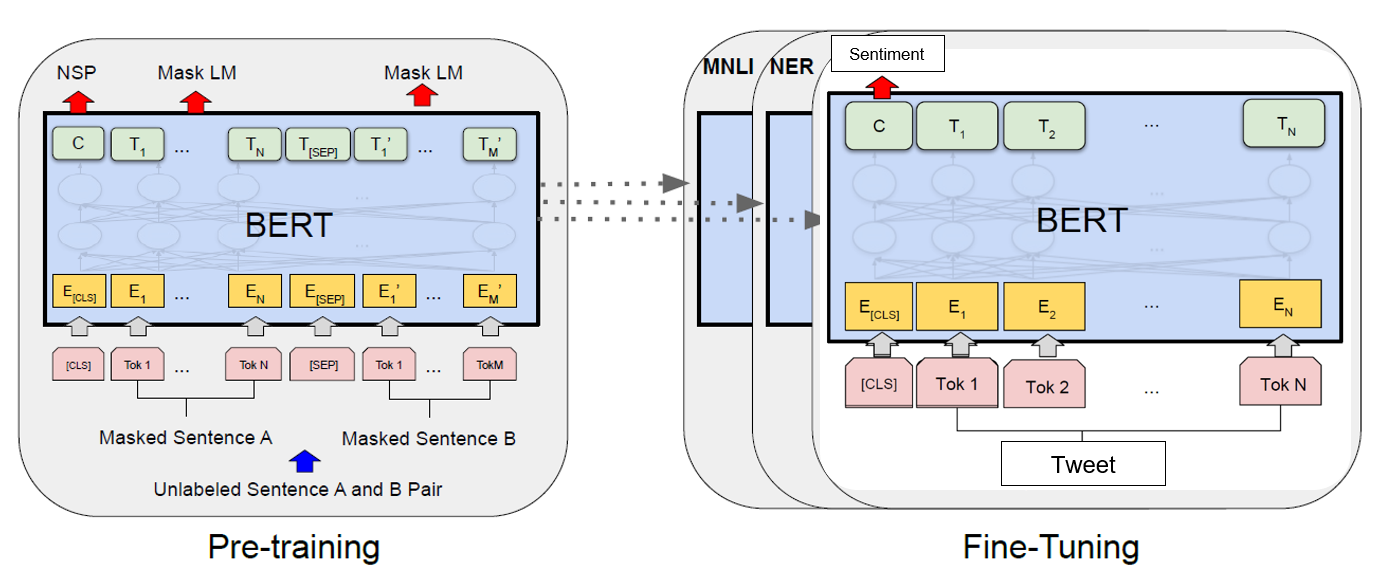}
    \caption{Architecture of a BERT model with the two training phases illustrated by \cayNP{devlin2018bert}}
    \label{BERT}
\end{figure}

Since text is available almost everywhere, the opportunities that bring NLP are really massive. The release in 2019 of latest text mining models brings new opportunities of services. Among those models, BERT (see section \ref{BERT} page \pageref{BERT} for more details) models\footnote{Initial BERT and all its extension} can for instance be used for direct information extraction through Question Answering (Q\&A). In the original paper presenting BERT \cay{devlin2018bert}, one of the task it was fine-tuned for was Q\&A, using the Stanford Question Answering Dataset (SQUAD) specifically developed for this task (see details about BERT section \ref{bertsection} page \pageref{bertsection}). The SQUAD consists of various English texts. With each document of the data set comes questions that are answered with an extract from the document. 

\begin{figure}[H]
\centering
\includegraphics[width=15cm]{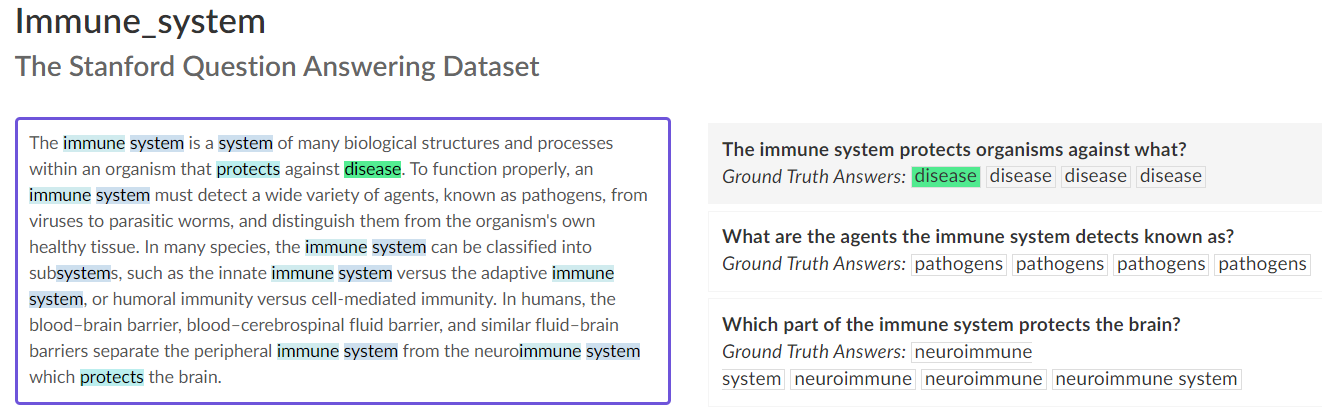}
\caption{Extract of the SQUAD taken from the source website \url{https://rajpurkar.github.io/SQuAD-explorer/} }
\label{SQUAD}
\end{figure}

One application of Question Answering for insurance companies is information retrieval from medical surveys. This example was developed in our library demonstrating Q\&A on different types of texts. After the release of the first BERT model, many engineers came up with their own versions of the model by adapting the original architecture. In this example we use a french version called CamemBERT. The version we use is one that was fine-tuned by French company Illuin's engineers on a dataset for Question Answering they built on their own called FQUAD (French Question Answering Dataset) \cay{dhoffschmidt2020fquad}. 

\begin{figure}[H]
\centering
\includegraphics[width=15cm]{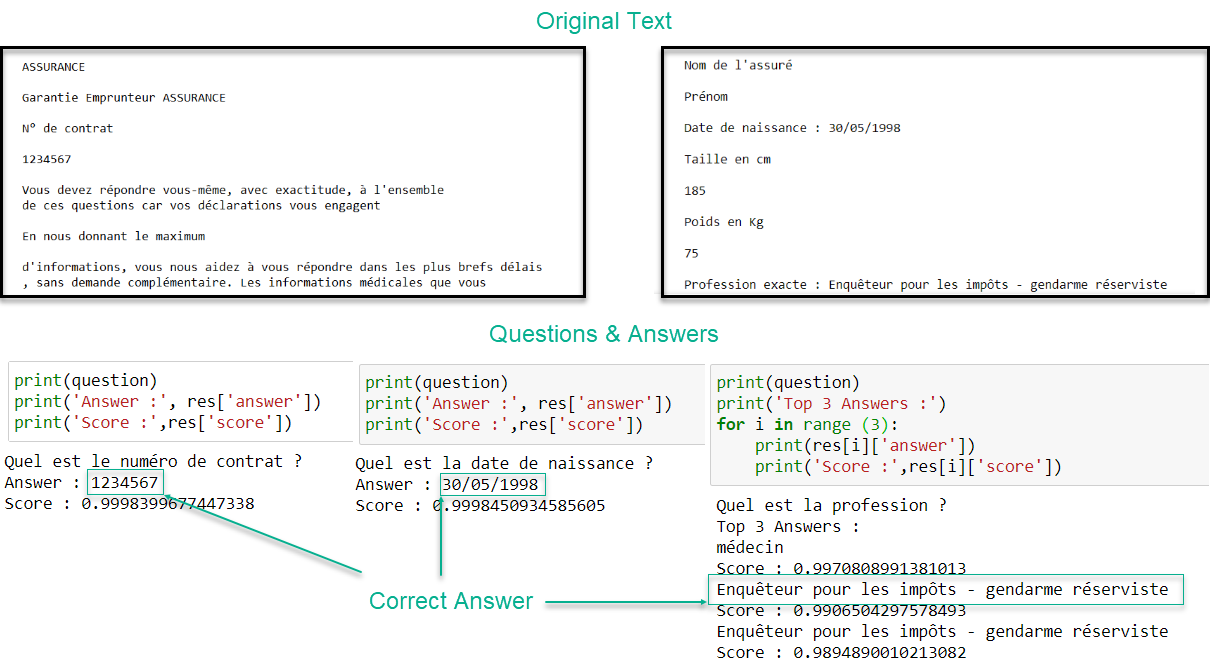}
\caption{Question Answering on a fake medical survey that can be used for automatic de-identification. If the answer given by the model is not correct, we display the top k answers to examine the other candidate answers.}
\label{FQUAD}
\end{figure}

The model proved to be efficient on simple and explicit questions, however, as shown figure \ref{FQUAD}, when asked for the profession of the fake person, it gives out the wrong answer (\textit{médecin}, which translates to doctor). This can be understood as the word \textit{médecin} appears a lot throughout the survey. A solution to have better results and use the model for more complex medical and insurance related questions could be to fine-tune the model one more time, on a dataset of medical surveys and insurance documents.

\section{Some history to understand evolution of NLP and new opportunities}

Even if NLP is more and more popular nowadays, it is not a so young discipline. The first automatic extraction of information from text dates back to the 1970s with the work of \cayNP{DEJONG1979251}. Most often referred to as \textit{Natural Language Processing} (NLP), semantic analyzes were subsequently developed in the 1990s \cay{grishman-sundheim-1996-message} with the creation of the \textit{Message Understanding Conference} (MUC). If research has intensified since then, the MUC initially proposes extracting linguistic information by using templates in which fields can be filled. Very useful to collect information in administration, once extracted, text can automatically fill other templates according to the information extracted from the values indicated by the user (e.g. generation of contract, certificate, etc.). In this approach, the structure of the response is assumed to be known. So in some fields, a simple common name is expected while in others, a sentence or adjectives can be filled. The extraction of information from templates therefore presupposes that we know in advance the structure of the information that will have to be extracted (for example a postal code in an address, an entity name or even a relationship). This approach is thus very limited in many situations. Less restrictive alternatives have appeared in order to be able to extract information from free text. Particularly favored by the development of the internet and the growth in the number of websites full of textual information \cay{Aggarwal2012}, the researchers are initially interested in using semi-structured text in HTML format \cay{Banko2007}. We therefore distinguish two major tasks which form the basis of the NLP \cay{Jiang2012}:

\begin{itemize}
    \item Recognition of the name of an entity designated most of the time by the acronym NER (\textit{Name Entity Recognition}): these techniques are concerned with extracting, from a text, the value of a common or proper name but also extends to the recognition of metric values (date, price, etc.).
    \item The extraction of relations, referring to techniques trying to determine the logical links between the different words of a text. From a raw or semi-structured text, the challenge is to find the grammatical relationships between the terms.
\end{itemize}

\subsection{Name Entity Recognition}
\label{NER}
The easiest way to implement name entity recognition is to use rules. When a pattern is detected, a rule is applied. The use of regular expressions \cay{regex} makes it possible to extract textual information which follows a particular format. For example, "Mr /. (. *)" allows you to  look for all the texts starting with "Mr" and to be interested in the group which follows it which is supposed to represent the name of the person. More precise rules allow to specify the authorized character values with which the pattern is supposed to correspond. Rules by regular expressions are nowadays commonly used in industry and are integrated in many products. \\

Other approaches like statistical learning approaches exist. In this context, the general idea is to be able to assign to any word (making up one document), one category indicating whether the word represents a particular entity to be extracted or not. In other words, statistical learning addresses the question of the recognition of entities through the form of a classification problem. As indicated by \cayNP{Jiang2012}, in this situation, the labels associated with the different words generally follow a convention called BIO \cay{BIO}.

\begin{figure}[H]
  \centering
  \includegraphics[scale=0.8]{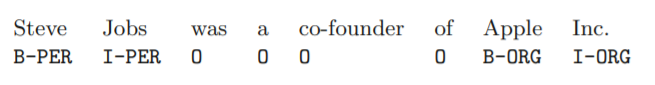}
  \caption{Illustration of categories following BIO convention}
  \label{bio}
\end{figure}

\cayNP{HMMRabiner} and \cayNP{Nymble} were pioneers in the study of NER algorithms by looking at the probability that one word is from a given category knowing the other words of the document. \cayNP{Nymble}  suggests in particular a supervised learning approach in order to determine the role of each word in a document. This approach is very similar to many implementations of text mining models.

\subsection{The study of relationships}

Other methods consist in defining relations between the extracted entities. For example, in the sentence "\textit{SCOR Data Analytics team has been deploying services using text mining models}", extracting entities would lead to defining "\textit{SCOR Data Analytics}" as a proper noun and "\textit{services}" as a direct object. However, the extraction of entity name does not make it possible to infer that "\textit{SCOR Data Analytics team}" is the team which deploys the "\textit{service}". This relationship study task can be approached as a classification problem between two co-occurring entities. \cayNP{JiangZhai} introduce a systematic approach by representing the relationship between the entities by a directed graph.

\begin{figure}[H]
  \centering
  \includegraphics[scale=0.8]{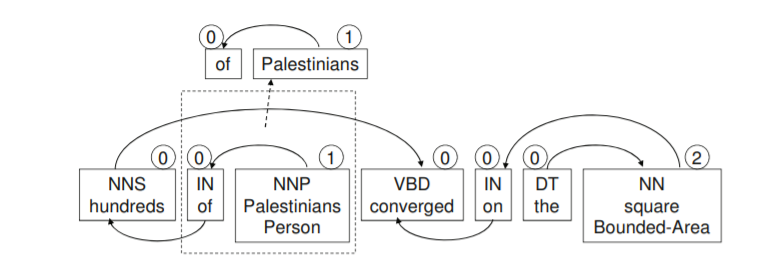}
  \caption{Example of dependence tree introduced by \cayNP{JiangZhai}}
  \label{bio}
\end{figure}

Following the idea of graphical representation of the connections between entities and of being able to extract subgroups of connected entities, kernel methods based on trees are proposed by \cayNP{Zelenko}. These methods will be extended in particular with the application of convolutional trees \cay{Zhang2006}. Convolutions allow a representation of a vector in a new space. In a convolutional tree, each dimension of the vector can then be associated with a subset of the tree and \cayNP{JiangZhai} then exposes examples of the use of this method. \\

\begin{figure}[H]
\centering
    \includegraphics[scale=0.5]{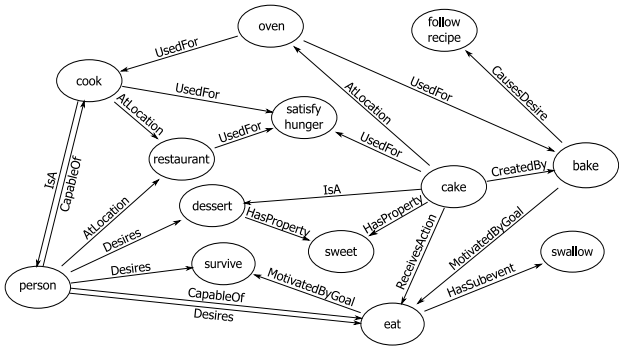}
    \caption{ConceptNet}
    \label{ConceptNet}
\end{figure}

If the methods have evolved a lot since the early 2000s, the desire to establish relationships between words - and thus adopt a grammatical approach to semantic analysis - remains valid. More recently, \cayNP{speer2017conceptnet} offer a general representation of writing and in particular of language, also using graph theory. The result, called \textit{ConceptNet} \footnote{\url{http://conceptnet.io/}} is thus a logical graph which connects words or sentences. Its construction is based on data provided by experts and others from questionnaires. The motivation is thus to provide a new representation of the language and improve the understanding of the language \cay{speer2017conceptnet}. The constitution of this graph has been greatly favored by the development of parallel computing techniques and the amount of data available. Thus the nodes (or the words) are put in relation by connections like: "\textit{Is used for}", "\textit{Is capable of}", "\textit{Is a}". Figure \ref{ConceptNet} illustrates an extract.

\subsection{The statistical approach and the use of Machine Learning}

If the extraction of information from free text makes it possible to respond to the detection of entity names or the study of the relationships between terms, the use of text can also be done in order to solve regression problems or more standard classification \cay{Aggarwal2}. Rather than considering words as entities related by attributes, one possibility is to simply consider words as occurrences of a random variable. Similar to the method introduced in the statistical approach allowing the recognition of entities, the documents are then described by statistics \cay{Nenkova} without however taking their order into account. The idea is to consider that words frequently appearing together in a document reveal information. \\

Thus, a first statistical representation of the text consists in representing each document by counting words occurrences according to a dictionary. The resulting counting vectors dimensions then represent a word from the dictionary (see section \ref{NLPdetails} page \pageref{TEsection} for more illustrations). In this representation, called Bag of Words (BoW), the idea consists in simply summing the canonical vectors associated with the words making up the sentence. In practice, the dictionary is rarely defined as the entirety of the words of a language. It is made up of the words observed in the corpus of documents used in the study.

This representation in Bag of Words, though it is simple, has the disadvantage of not being sparse. Indeed, in a sentence, we generally use only few words among those existing in the dictionary. On the other hand, articles, linking words or common terms (called \textit{stopwords}) are often over-represented by this method. To cope with this last drawback, a heuristic consists in fixing a certain list of unwanted words and removing them from the dictionary. \\

In a specific set of documents -called corpus, the frequency of certain words may also be higher than in others. Thus, an alternative to the representation in a bag of words is one which no longer simply counts the occurrence of a word in the document but also takes account of its appearances in the corpus.

Text mining and NLP research over the past 10 years has attempted to overcome the various limitations of early historical methods. The past five years have disrupted the state-of-the art text mining models by using deep learning \cay{mikolov2013efficient} \cay{vaswani2017attention} \cay{devlin2018bert}. The most recent models are able with high precision to capture not only words interaction but are also able to adapt the focus on specific words according to the context of the sentence (cf section \ref{attention}). Those most recent NLP models provide new opportunities in client relationship but also improvement in document analysis.

The following sections help in understanding how to use the different machine learning models with text data and apply them to industrial cases. We particularly focus on the latest NLP models but reader might refer to \cay{1023184} for a reminder on more traditional machine learning models mentioned in this article that can be combined with NLP models.

\section*{NLP in practice: from data processing to production}

\begin{figure}[H]
    \centering
    \includegraphics[width=8cm]{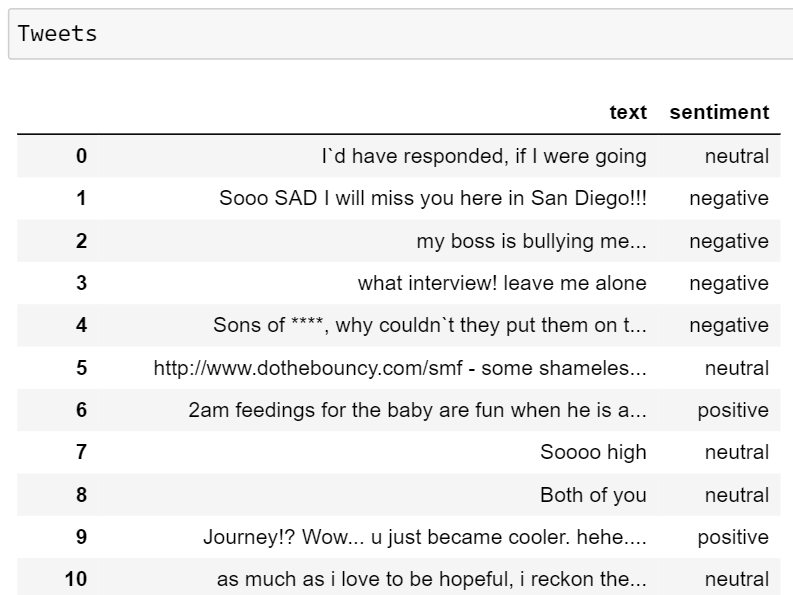}
    \caption{Extract of Kaggle's Tweet for Sentiment Extraction Dataset.}
    \label{Jupyter0}
\end{figure}

In practice, text manipulation follows standard pipelines in order to answer the initial business issue motivating the usage of NLP.
In this section, we illustrate the different steps that are applied in practice when manipulating text data. Some illustrations are given in Python and based on the Twitter’s Tweet for Sentiment Extraction Dataset (cf Figure \ref{Jupyter0}) or on other use cases to illustrate some specific steps. 
This section provides a high level overview of techniques and are addressed to non-expert readers. Deeper explanations for each step of the process are available in section \ref{NLPdetails} page \pageref{NLPdetails}.

\section{Text mining illustrated through use cases} 

If every project has its own specificity (underwriting, claims analysis, fraud detection, etc.), there are some steps that are common when it comes to manipulating text. For this reason, a Python library has been developed by SCOR Data Analytics team to facilitate the use of text mining\footnote{Please contact authors for more information}. The developed library reflects each of standard step to process text. It embeds some open source libraries and adds some customized features that help dealing with different languages and make usage uniform.\\

To decrease the learning curve of some functionalities, a few notebooks have been created, involving some preprocessing steps (implemented in the module `preprocessing` as illustrated in Figures below). For instance, the notebook entitled \textit{[TUTO]Data\_analysis}, presents and end-to-end study applied on a dataset of tweets along with a sentiment score (negative, neutral or positive), from the Tweet Sentiment Extraction\footnote{\url{https://www.kaggle.com/c/tweet-sentiment-extraction/data?select=test.csv}} Kaggle competition.

In the following paragraphs, this example is taken as an illustration for the different transformations that allow text manipulation by algorithms.

\section{Data is never clean: the preprocessing} 
\label{preprocessing}
Some words are not useful in term of predictive power because they are too frequent, therefore not very discriminant. The goal of the preprocessing is to prepare text representation so it can feed a parametric models (e.g. linear regression) by cleaning the data. Indeed, all strings must have a similar form and unnecessary elements must be removed from our inputs since they can introduce noise. The most essential preprocessing steps are:
\begin{itemize}
    \item Loading the text: It is common, when dealing with specific characters like accents to face encoding issues. This part aims to solve text representation and encoding problems using the "unicodedata" library. Encoding conversion from Latin1 to UTF8 (e.g. when dealing with mix of French and English words) can be applied before processing the text.
    \item Lowercasing: It is the operation of converting all letters to lowercase. This will get rid of any ambiguity between lower-case and upper-case letters. This step can easily be applied with built-in methods associated to string objects (".lower()" for instance in Python).
    \item Stopwords removal: When communicating, humans must use words such as “the”, “and” or “a” for sentences to make sense, however these words do not carry any meaning and are not useful for data analysis. Getting rid of these words called stopwords is a preprocessing step that allows focusing on words that matter. Each language possesses its own list of stopwords that can be adjusted according to the problem. The first idea of stopwords appeared in an article by \cayNP{doi:10.1002/asi.5090110403}. Lists of predefined stopwords for many languages are available in NLTK\footnote{\url{https://www.nltk.org/}}. 
    \item Special characters processing: an attention is brought to special characters. Indeed, while some of them are not necessary, others are needed for sentiment analysis. For instance, an exclamation point in a tweet can express an emphasis on its sentiment. People tend to increase the number of exclamation points when dealing with strong sentiments. Therefore, punctuation characters can be kept or treated as stopwords. The "re" library is used to remove duplicated punctuation symbols as well as other superficial pieces of text like URLs or Twitter nametags thanks to regular expressions (see the beginning of section \ref{NER} page \pageref{NER} for more information about regular expressions).
    \item Tokenization: In order to study a text, the data must be split into sentences and possibly into words. This operation of splitting a text into smaller parts is called tokenization. A lot of methods cover tokenization, from Stemming \cay{M.F.Porter1980} to WordPiece \cay{schuster2012japanese}. See section \ref{lowercasingtokenization} page \pageref{lowercasingtokenization} for further details. In practice, libraries differ according to language for stemming (e.g. Jieba for Chinese, Treetagger for French or NLTK for English). Recently, models like BERT detailed in section \ref{NLPdetails} page \pageref{BERT} can deal with multi-language data with a unique model.
\end{itemize}

As preprocessing is utterly basic, it is only natural to standardize it using the previously cited libraries within simple functions, which is what we have done to simplify preprocessing in text mining projects and really focus on the business issue. Thus, the preprocessing steps are adapted to any language using latin characters (French, English, German, etc.) as well as Chinese, and can be conveniently used by calling functions from our own library. We invite readers to contact authors for any sharing.

\begin{figure}[H]
    \centering
    \includegraphics[width=15cm]{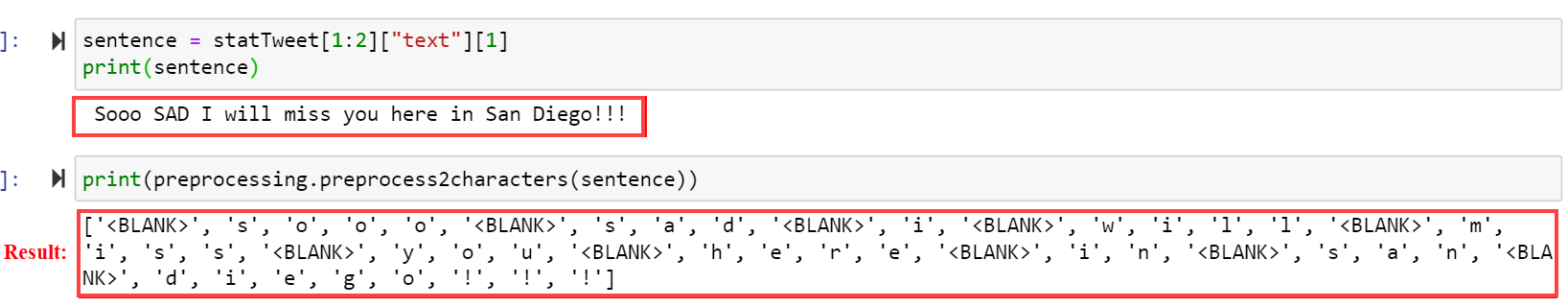}
    \caption{Character level preprocessing.}
    \label{Jupyter2}
\end{figure}

Figure \ref{Jupyter1} and \ref{Jupyter2} illustrate some results of preprocessing raw tweets from the Tweet for Sentiment Extraction Dataset using SCOR library:

\begin{figure}[H]
    \centering
    \includegraphics[width=15cm]{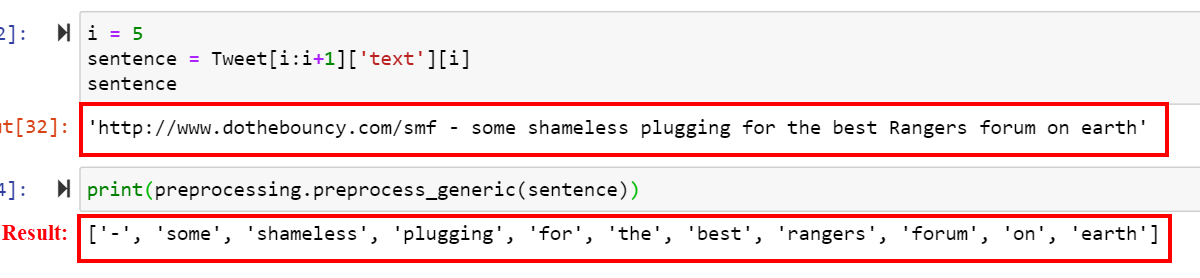}
    \caption{Word level preprocessing.}
    \label{Jupyter1}
\end{figure}

Textual data may have spelling or grammar mistakes. In this situation, Character Level preprocessing can be useful. It is particularly necessary when working on tweets.

In practice, the preprocessing is usually adapted to the considered problem but also to the language. For instance, Chinese characters need to be split in a specific manner as shown below, because the fact that words are not separable by single character prevents from using the same preprocessing method as in latin language.

\begin{figure}[H]
    \centering
    \includegraphics[width=15cm]{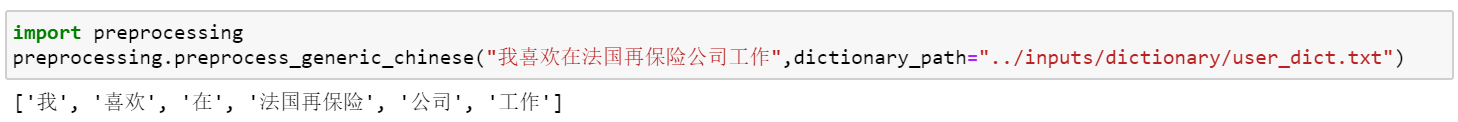}
    \caption{Chinese data preprocessing.}
    \label{Jupyter5}
\end{figure}

\section{Converting text to numerical vectors: Text Embedding}
\label{textembedding}

Once the data is clean, the next step is transforming textual data into numerical vectors, through an operation that is called text embedding. Indeed, a computer cannot understand raw words and learn from them. It can only work with numbers. Therefore, an appropriate representation with numbers must be adopted. 

First a dictionary is built with the available vocabulary. Then, given a sentence from the dataset, each word/character is referenced to as a token, according to its rank in the dictionary. This process is also called tokenization, however it is not the same process as defined in section \ref{preprocessing} page \pageref{preprocessing}. The two definitions are sometimes wrongly mixed-up. In order to avoid any confusion, we will talk about vocabulary tokenization when mentioning lexical indexation.  




\begin{figure}[H]
    \centering
    \includegraphics[width=15cm]{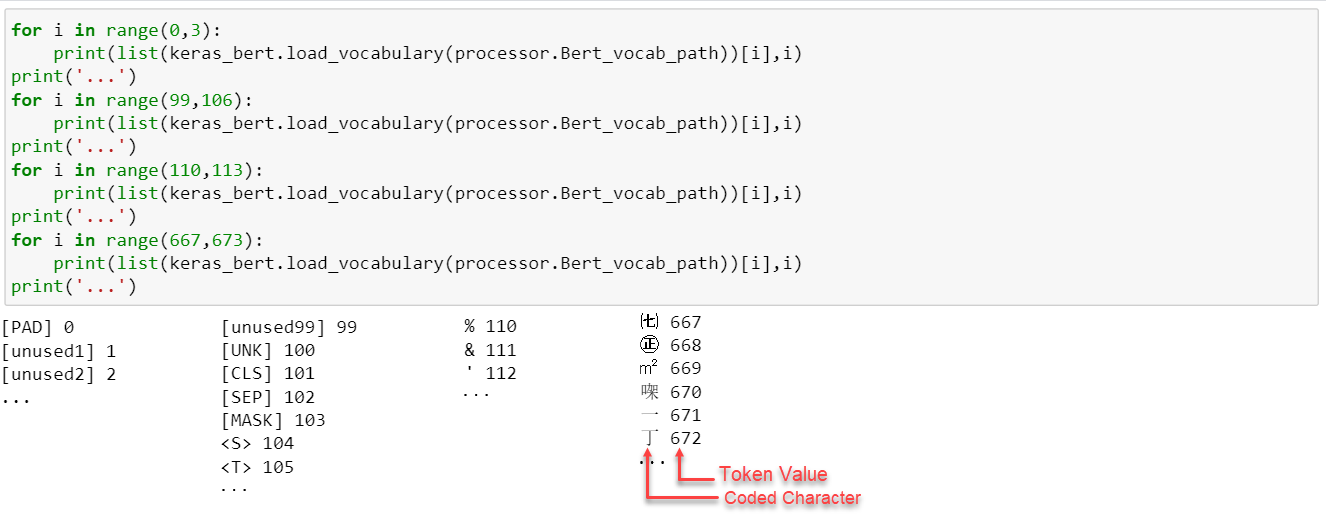}
    \caption{Dictionary used for vocabulary tokenization.}
    \label{Jupyter3}
\end{figure}
 As shown below, this method transforms textual data into a numerical vector.
\begin{figure}[H]
    \centering
    \includegraphics[width=15cm]{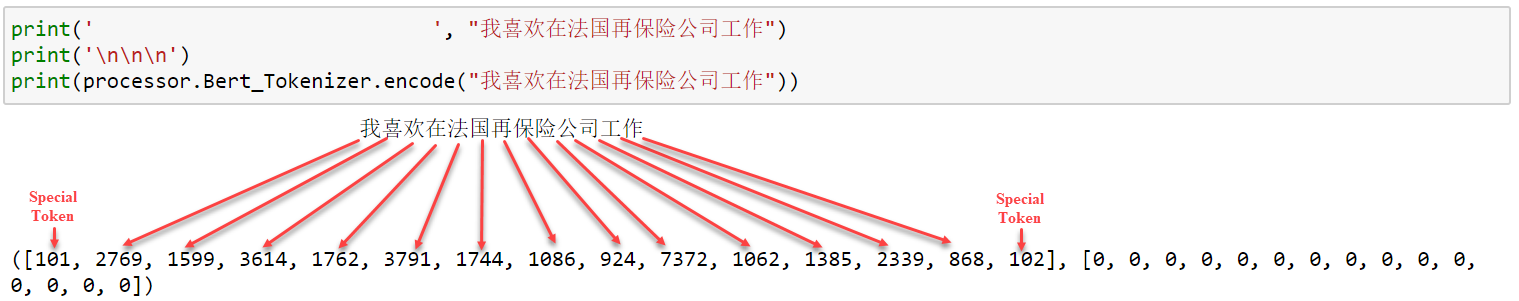}
    \caption{Vocabulary tokenization of a Chinese sentence.}
    \label{Jupyter4}
\end{figure}

In the example shown figure \ref{Jupyter4}, the tokenized vector is completed with "0". This is done because every tokenized vectors are required to have the same size. Usually, it is set to the size of the longest availabe tokenized sentence.

After tokenizing words into indexes of a dictionary, other transformations are applied, taking factors like context or semantics into consideration. Several methods cover these transformation processes. One of the most famous methodologies is "Word2Vec" \cay{mikolov2013efficient} which is detailed in section \ref{TEsection} page \pageref{TEsection} with other alternative models. Text embedding is the combination of vocabulary tokenization and other transformations. It is common in the industry to leverage on pre-trained models to face lack of data in training set or to reduce computation times. The most popular pre-trained NLP embedding models are details in section \ref{TEsection} as well. 


\section{Feature engineering.}

To come back on the previous example on Sentiment Analysis, it is also common and useful to create new features that could bring information. For instance, length of one's comment, the count of some specific punctuation, etc. It is quite difficult to provide an exhaustive list of feature engineering that can be applied to text since it is usually adapted to a specific issue.\\

Some tricks can however be useful in the learning phase of an NLP model. Among them is the data augmentation that can be done thanks to translators. Coupling translation algorithms can help in enriching your data set and keep the meaning of your text. For instance, translating from English to French then French to English through another translator will return a similar sentence but not exactly the same. The longer are the paragraph you translate, the more useful is that trick.\\

Other feature engineering methods like special character interpretation can help in pre-processing your text. When you deal with tweeter for instance or conversational text, interpreting emoji might provide relevant information. Converting them into words before manipulating the full text can sometimes enhance the information quality. \\

For more specific use cases, like in life insurance, injecting some knowledge around medical vocabularies can also help in gathering similar words (e.g. diseases).

\section{Modelling with NLP.}

Once an input is embedded in the form of a numerical vector, a model has to be trained to deal with the main task.

\begin{figure}[H]
    \centering
    \includegraphics[width=15cm]{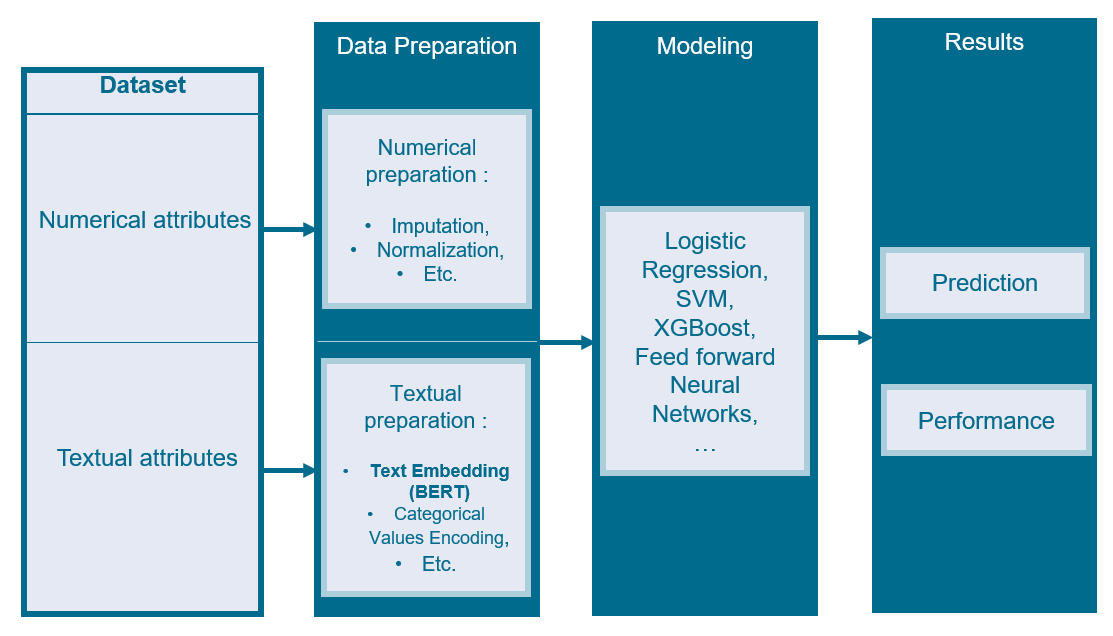}
    \caption{Usual strategy when dealing with common datasets where NLP is not the main task, but still required for efficient text embedding.}
    \label{Practical}
\end{figure}

In the case of tweet sentiment extraction, after applying an efficient text embedding, the modeling part would address the classification problem. To each tweet, a sentiment has to be correctly attributed. Normally, the data preparation and modeling part are done separately. However, when the data is only textual, BERT can be used for both (see section \ref{bertsection} page \pageref{bertsection} for more details).\\

In practice it is also common to merge numerical embedding vectors with regular tabular data in order to enrich the available features. For instance, in traditional insurance data, for each policy, the description could be enriched with columns that contain text information. Thus numerical added features can then feed the training of a specific model.\\

After training, the model requires performance evaluation, which can be done through the choice of suitable metrics (see section \ref{metrics}) and rigorous study of interpretability. 

\section{How to test and analyze a text mining model?}



NLP models are tools to enhance business products and help in solving issues that tabular data alone cannot solve. In more traditional predictive analytics only based on tabular data and statistical models, the interpretation of the model usually takes large part of the model analysis. Since models are used to better understand phenomenons (and then better anticipate them), matching the expert judgements with the model outcome is key to ensure relevance of the predictive model. It ensures the control but also the interpretation. 

In machine learning, there is a trade-off between model complexity and model interpretability. Complex machine learning models e.g. deep learning perform better than interpretable models. However, complexity can sometimes lead to black box effects. Research has been active for the past few years in providing additional instruments to open those black box models \cay{delcaillau2020interpretabilit}.

For simple vector embeddings like BoW, Tf-Idf (cf section \ref{TEsection} page \pageref{TEsection} for more details), it is easy to use LIME or SHAP on top of the machine learning models to get some insights of the results. LIME stands for Local interpretable model-agnostic explanations. The idea behind this is to have a local linear approximation of the complex model. Alternatively, SHAP (SHapley Additive exPlanations) unifies all available frameworks for interpreting predictions and it is based on the game theory concept. LIME and SHAP models are surrogate models that model the changes in the prediction. However it is important to note that these two methods are univariate and do not take into account the correlations between variables. Many other interpretation tools can then be added in complement to the embedding methodology and the model used to achieve a specific issue. The following figures illustrate SHAP and LIME’s explanation for a positive tweet\footnotemark. 

\footnotetext{Two tutorial notebooks are available in our library, \textit{[VisualizationNotebook] Explain model outputs LIME.ipynb} and \textit{[VisualizationNotebook] Explain model outputs SHAP.ipynb}.}

\begin{figure}[H]
    \centering
    \includegraphics[width=15cm]{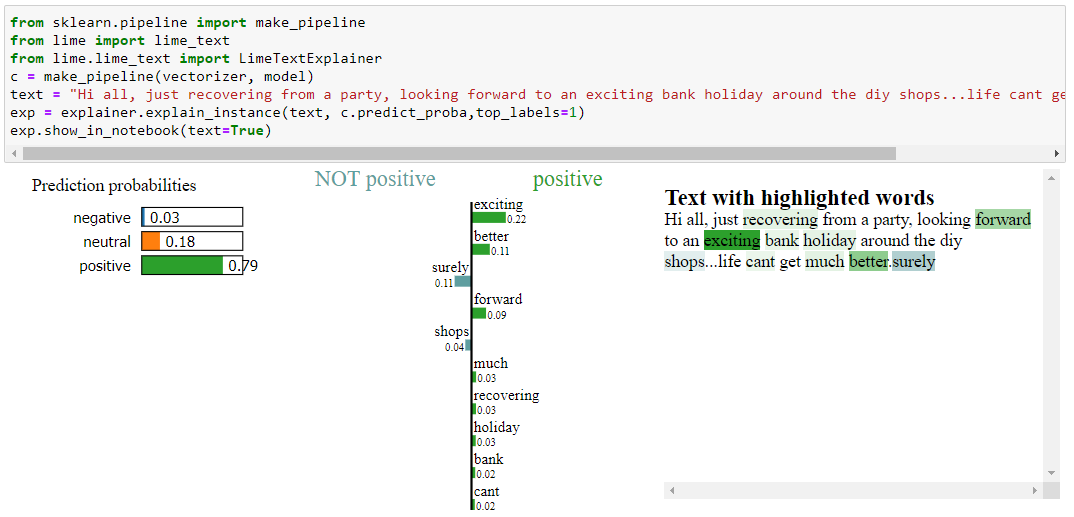}
    \caption{Results analysis with LIME explainer in Tweet Sentiments Predictions}
    \label{Jupyter6_lime}
\end{figure}

\begin{figure}[H]
    \centering
    \includegraphics[width=15cm]{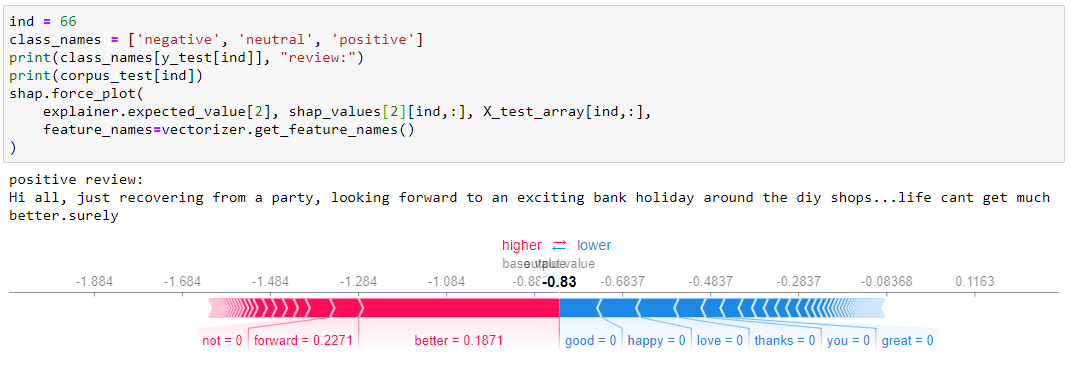}
    
    \label{Jupyter7_shap}
    \floatfoot{The above explanation shows features each contributing to push the model output from the base value (the average model output over the training dataset we passed) to the model output. Features pushing the prediction higher are shown in red, those pushing the prediction lower are in blue (these force plots are introduced in the Nature BME paper \cay{lundberg2018explainable})}
    \caption{Results analysis with SHAP explainer in Tweet Sentiments Predictions}
\end{figure}

With more complex word representations such as BERT, algorithm will have a better accuracy, but there is less possibility to explain each output individually. However, it is still possible to directly analyze the prediction and explain it from text. For instance, one can do a word frequency analysis and get a quick look at a word cloud per classes. Whatever the model, we can always use common sense.

\begin{figure}[H]
    \centering
    \includegraphics[width=10cm]{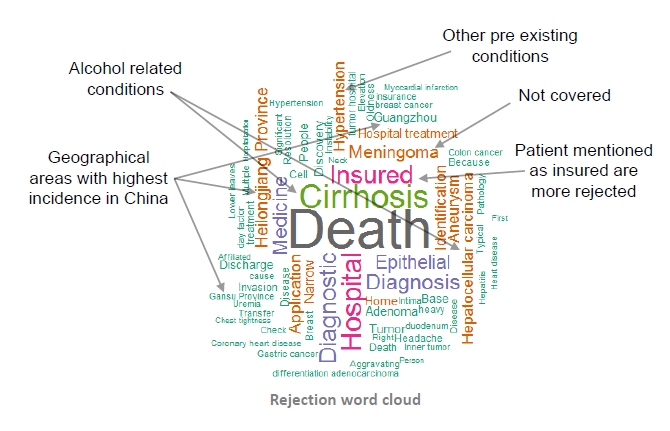}
    \caption{Word cloud of the class rejection}
    \label{Wordcloud}
\end{figure}

\section*{Further description of NLP}

\label{NLPdetails}

\section{Deeper understanding of text preprocessing}
\subsection{From lowercasing to tokenization}
\label{lowercasingtokenization}
As explained in the previous chapter, words need to be treated before being converted to numbers and processed by a model. In order to do this, after a simple lower casing, a sentence is tokenized into smaller parts (often into words using a well thought regular expression), and the words are regrouped in a list. In the example of sentiment analysis, some punctuation characters like "!" are useful, as they can express a sentimental emphasis, therefore they are kept during this step. However, URLs are removed during this phase, as they provide no useful data for this specific analysis. 

\begin{figure}[H]
    \centering
    \includegraphics[width=14cm]{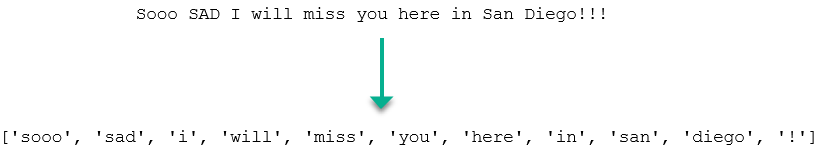}
    \caption{Simple tokenization of a tweet.}
    \label{tokeniz}
\end{figure}

There are different methods of tokenization. The one illustrated figure \ref{tokeniz} is tokenization of a sentence into words. However, words can also be divided into smaller parts. The point of tokenizing words is to find similarity within words with the same roots. In order to understand this, two popular methods called Stemming and Wordpiece tokenization can be studied.

\subsubsection{Stemming}
Grammar rules often demand some suffixes to be added to words for a sentence to make sense, however those suffixes are not imperatives when it comes to understanding the meaning of a word. Stemming is a preprocessing step that gets rid of the unneeded parts of a word while keeping its root, also called the "stem".

\begin{figure}[H]
    \centering
    \includegraphics[width=12cm]{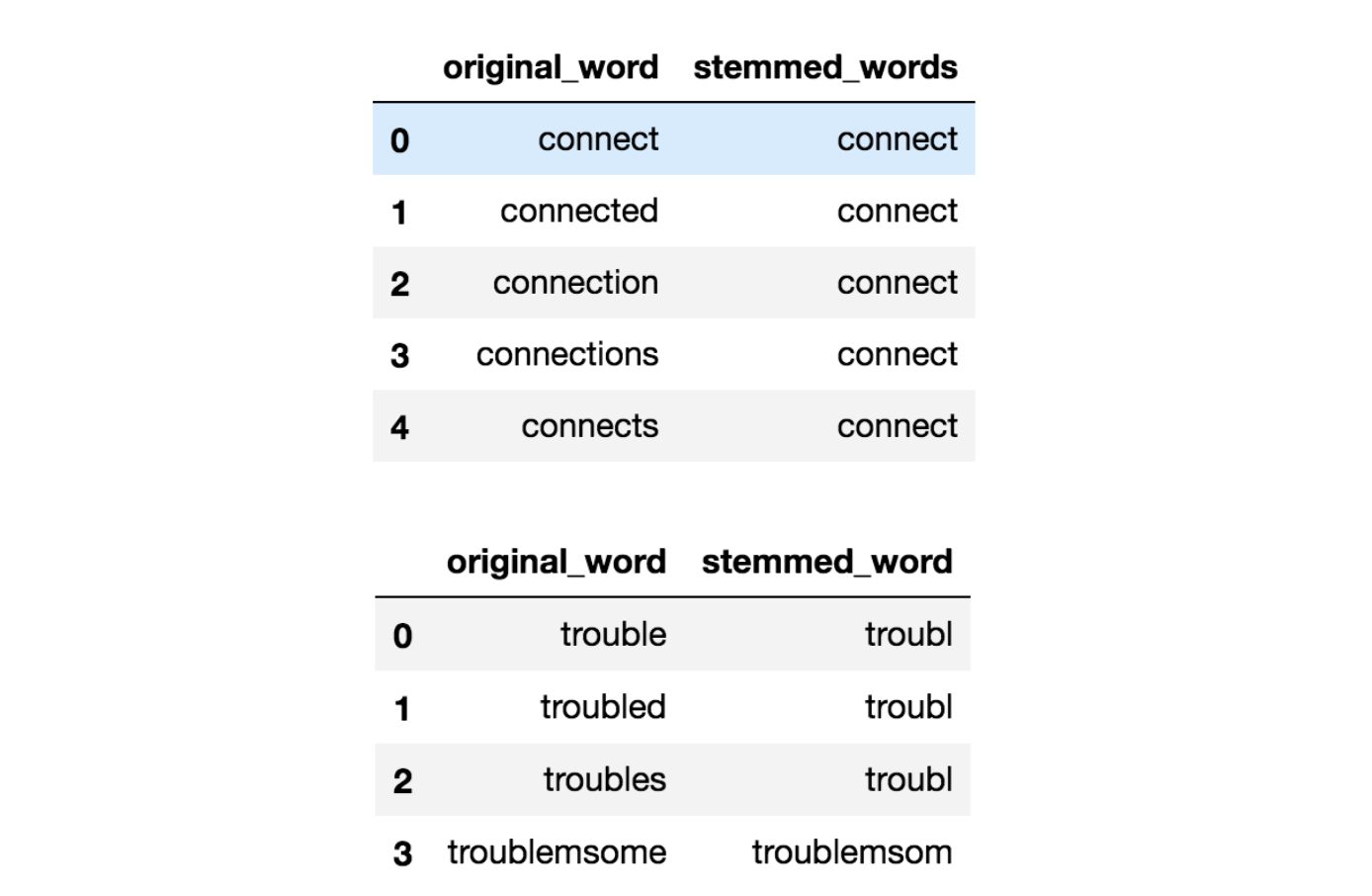}
    \caption{An example of stemming. Illustration by \cayNP{kdnuggets}.}
    \label{stemming}
\end{figure}

In order to achieve such a result, two algorithms are mainly used: the Porter Stemming Algorithm, introduced by \cayNP{M.F.Porter1980} and the Lancaster Stemming Algorithm (also known as Paice/Husk Stemming Algorithm) developed at Lancaster University by \cayNP{Lancaster2005}. 
The Porter Stemming Algorithm is made of 5 steps or set of rules based on the structure of words that are applied in a specific order. These rules mainly focus on the combination of specific vowels and consonants.
The Lancaster Stemming Algorithm is also based on a set of rules defined to remove suffixes, but though it is less complex and faster, this algorithm is iterative and can cause over-stemming, returning stems that are difficult to understand.

\subsubsection{Lemmatization}
While stemming rids a word of suffixes and prefixes and only keeps its root, lemmatization is an algorithm that replaces a word by its most basic form, also called "lemma". A lemma can be an infinitive form, a noun, an adjective, etc, while a stem often means nothing. This is useful because in some languages, words with different meanings can have the same stem. Therefore, lemmas aim to convey an idea with an actual word instead of a basic stem with no definite meaning, correcting the problem of having two different concepts with a same root. Lemmatization thus requires a morphological analysis of the word and the existence of a detailed dictionary for the algorithm to work on, making it more complex to implement than stemming.

\begin{figure}[H]
    \centering
    \includegraphics[width=8cm]{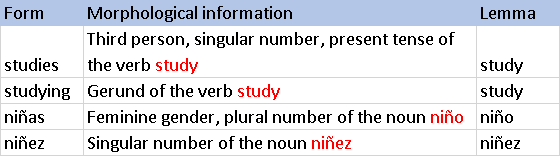}
    \caption{An example of lemmatization by \cayNP{bitext}.}
    \label{lemmatization}
\end{figure}

\subsubsection{WordPiece tokenization}
\label{WordPiece}
While stemming and lemmatization gets completely rid of suffixes, the WordPiece method  \cay{wu2016googles} breaks words into smaller parts by keeping suffixes using a machine learning model. In the original paper, the WorldPiece model is trained on a vocabulary from 8K to 32K words. The purpose of this training is to learn how to tokenize sentences and words efficiently. 

\begin{figure}[H]
    \centering
    \includegraphics[width=8cm]{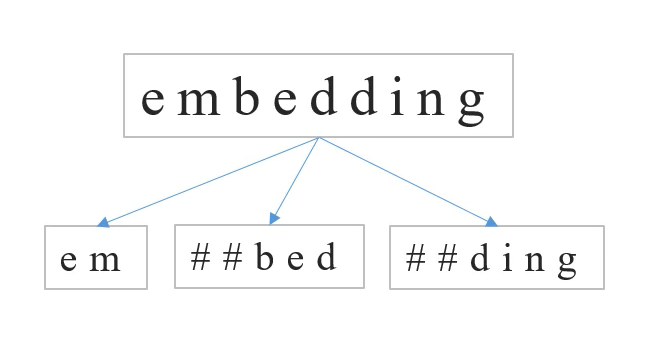}
    \caption{WordPiece tokenization on the word "embedding". Illustration by \cayNP{ChrisMcCormick}.}
    \label{WP}
\end{figure}

The "\#\#" characters shown figure \ref{WP} are used to denote all WordPieces that do not start a word. Training the model to spilt words into WordPieces makes it able to perform on unknown data. For instance, if the model has to tokenize a word it hasn't been trained on, it will try to look for roots it has studied during training and produce a segmentation accordingly. 

\begin{figure}[H]
    \centering
    \includegraphics[width=10cm]{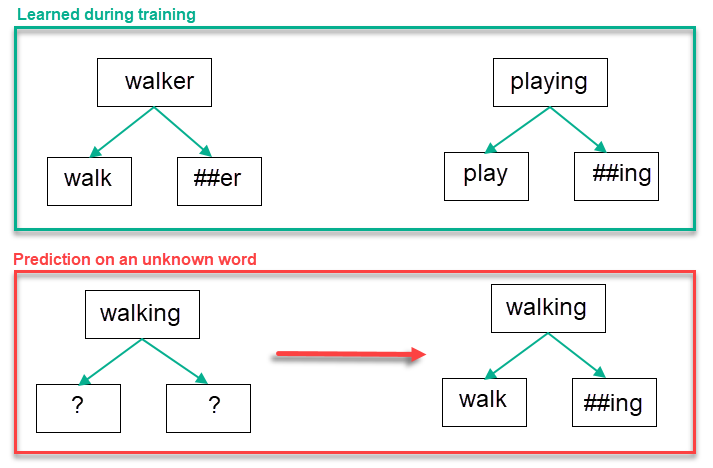}
    \caption{WordPiece tokenization of a word that is unknown to the model.}
    \label{WP2}
\end{figure}

\subsubsection{N-grams}
Sometimes, association of words can change the meaning or bring different information. If some embedding models as BERT (which is described further in this paper) already take into account the context, less recent approaches require some processing on the words (that can be stemmed). For instance, if we are predicting the sentiment of one sentence, the word "bad" would be expected to push the output toward the negative label. However, "not bad" would do the opposite. The traditional word decomposition then need to be enriched and we might need to consider "not bad" as a unique word. The construction of new "words" coming from the co-occurence of words is called "n-grams" with "n" standing for the number of words we would like to consider (in our exemple "not bad" is a "2-grams"). When the embedding model cannot capture the context, this processing aims to enrich the dictionary of words.

\subsection{Length of the text}

When dealing with long texts, NLP models tend to perform less efficiently, as they need to process every sentences. However, sometimes the sentiment is not distributed throughout the whole document, but can be expressed in a single sentence. For instance, when working on an article, the title can convey much more information than the actual article with context-setting and argumentation. Moreover, the longer the text, the more bias is introduced by stopwords, reinforcing the need to remove them. Therefore, the length of the text is a factor to keep in mind when dealing with NLP tasks.

\section{Deeper understanding of text embedding}
\label{TEsection}
After explaining the technical aspects of preprocessing raw words, let's have a closer look to the methods used to make words understandable by a computer. The different subsections below introduce the most popular embedding techniques that are used in Natural Language Processing.
\subsection{Bag Of Words (BoW) and Term Frequency-Inverse Document Frequency (Tf-Idf)}

The most fundamental method of text embedding is Bag Of Word first mentionned by \cayNP{doi:10.1080/00437956.1954.11659520}. In order to explain the latter rigorously, let us define some notations. 
\begin{itemize}
    \item Let $c\in\mathbb{R}^{n}$ be a vector of dimension $n \in \mathbb{N}$ called "character". It could be letter or punctuation or any unitary symbol. 
    \item We define a "segment" $w\in(\mathbb{R}^n)^{\mathbb{N}}$ as a finite sequence of "characters". It could be a word (if the "characters" are letters). 
    \item We define a "dictionary" as a subset $\mathcal{D}\subset{(\mathbb{R}^n)^{\mathbb{N}}}$
    \item We define a "document" $d$ $\in\mathcal{D}^{\mathbb{N}}$ as a finite sequence of "segments". 
    \begin{center}
        $d = (w_i)_{i \in \llbracket{1,n_d}\rrbracket}$ with $n_d$ the length of the document
    \end{center}
\end{itemize}

Given a document $d$, the BoW representation
displays the frequency of each word in each sentence of the document.

For instance, let’s study the document $d$ = [“He is a good boy”, “She is a good girl”, “Boys and girls are good”]. After preprocessing, $d$ becomes $d’$ = [“good boy”, “good girl”, “boy girl good”]. Now let's build a dictionary with the available words in $d'$, that we denote $\mathcal{D}$. $\mathcal{D} = $ ["good", "boy", "girl"]. Using this dictionary and counting the occurrence of each words of the dictionary in each sentence, we get the following Bag of Words representation for the set $d’$:

\begin{table}[ht]
\centering
\begin{tabular}{llll}
  \hline
\textbf{ } & \textbf{good} & \textbf{boy} &\textbf{girl} \\ 
  \hhline{====}
$d’[1] = $ "good boy" & 1 & 1 & 0 \\
$d’[2] = $ "good girl" & 1 & 0 & 1 \\
$d’[3] = $ "boy girl good" & 1 & 1 & 1 \\
\hline
\end{tabular}
\captionof{table}{Bag Of Words representation. Example by \cayNP{Krish}.
}\label{bow}
\end{table}

Though this representation is simple to understand, it gives an equal weight to every word and no semantic difference between words. However BoW can be improved by adding Term Frequency, first mentioned by \cayNP{5392697} and Inverse Document Frequency \cay{KAREN1972}. 
Let us define a word $w\in\mathbb{R}^{\mathbb{N}}$, and a set $d$ $\in(\mathcal{D})^{\mathbb{N}}$ of sentences. Term Frequency $tf(w,d)$ of $w$ can be defined like so :

\begin{center}
$tf(w,d) = \frac{\text{number of times the word }w  \text{ appears in } d}{\text{total number of words in } d}$
\end{center}

Let us denote $len(d)$ the number of sentences in the document $d$ and $df(w)$ the number of sentences in $d$ containing the word $w$. Inverse Document Frequency can be defined like so :

\begin{center}
$idf(w,d) = log(\frac{len(d)}{df(w)})$
\end{center}

By multiplying $tf(w,d)$ and $idf(w,d)$ of each words we obtain a new matrix that gives us semantic differences through weights that define the significance of a word in a sentence.
In the previous example, this will look like this :
\begin{table}[H]
\centering
\begin{tabular}{llll}
  \hline
\textbf{tf} & \textbf{good} & \textbf{boy} &\textbf{girl} \\ 
  \hhline{====}
$d’[1]$ & 0.5 & 0.5 & 0 \\
$d’[2]$ & 0.5 & 0 & 0.5 \\
$d’[3]$ & 0.33 & 0.33 & 0.33 \\
\hline
\end{tabular}
\qquad
\begin{tabular}{llll}
  \hline
\textbf{} & \textbf{good} & \textbf{boy} &\textbf{girl} \\ 
  \hhline{====}
\textbf{idf} & Log(3/3)=0 & Log(3/2) & Log(3/2) \\
\hline
\end{tabular}
\captionof{table}{$tf$ and $idf$ on the same example. \cayNP{Krish2}}\label{bow}
\end{table}

\begin{table}[H]
\centering
\begin{tabular}{llll}
  \hline
\textbf{Tf-Idf} & \textbf{good} & \textbf{boy} &\textbf{girl} \\ 
  \hhline{====}
$d’[1]$ & 0 & 0.5*Log(3/2) & 0 \\
$d’[2]$ & 0 & 0 & 0.5*Log(3/2) \\
$d’[3]$ & 0 & 0.33*Log(3/2) & 0.33*Log(3/2) \\
\hline
\end{tabular}
\captionof{table}{$Tf-Idf$on the same example. \cayNP{Krish2}}
\label{bow}
\end{table}

\subsection{Word2Vec}
\label{def}
The BoW method presents two main disadvantages. The first one is that it deals with very large vectors and the second one is that there is no notion of similarity between words, as they are treated as indices of a vocabulary set. The Tf-Idf adds weights to words, however there is no proper relations between words.  \cayNP{mikolov2013efficient} suggested a vectorial representation of words that would consider the context from which they are taken, thus introducing the Continuous Bag Of Word (CBoW), which aim is to predict a word based on its surrounding context and the Skip-gram model, which aim is to predict the surrounding context based on a single word. These two models are better known as Word2Vec, as the original code was named.

\begin{figure}[H]
    \centering
    \includegraphics[width=15cm]{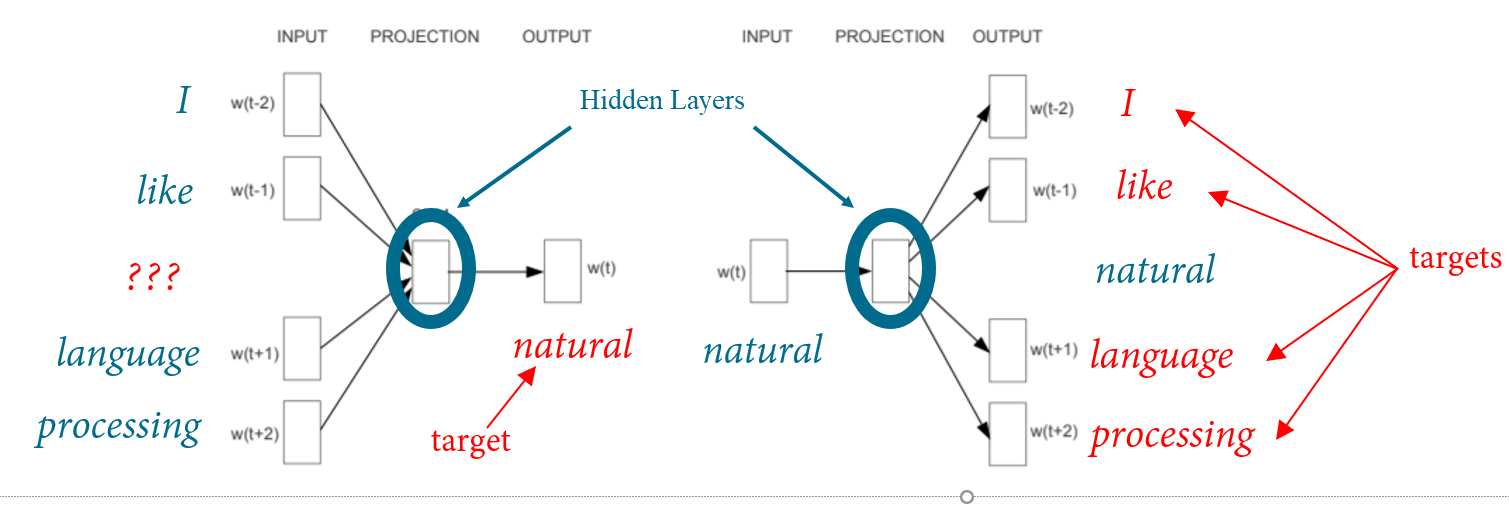}
    \caption{Architecture of the Continuous Bag of Word (left) and Skip-Gram (right) models. Illustration by \cayNP{W2Varchitecture}.}
    \label{CBOW}
\end{figure}

Let us have a segment $w$ of size $N$, which translates to a sentence of $N$ words $(w_1, ..., w_N)$. We define a context window $C$ of size $k$ that can be placed anywhere in the sentence. The aim of the CBoW model is to predict the center word of the context window $C$ by using the other words from the context window. This is achieved by training neural networks.

\begin{figure}[H]
    \centering
    \includegraphics[width=7.5cm]{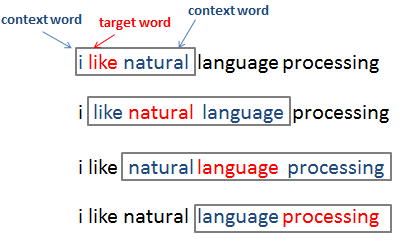}
    \caption{Illustration of a Context Window in Continuous Bag Of Words by \cayNP{W2V}. The aim of the training is to predict the red word using the blue words.}
    \label{CBOW2}
\end{figure}

The Skip-Gram algorithm works the opposite way, trying to predict a context window from the middle word (see red word figure \ref{CBOW2}). These two representations allow relations to be built between words. 

After the training we can extract from the hidden layer(circled in blue figure \ref{CBOW})  an ‘embedding layer’ or ‘embedding matrix’.
This matrix can then be applied to any context window and similar context windows will return similar words, creating relations between contexts and words. 
\begin{figure}[H]
    \centering
    \includegraphics[width=12cm]{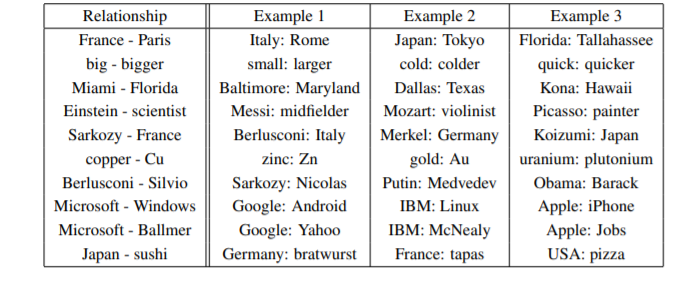}
    \caption{Examples of relations built by a Skip-Gram model from the original paper by \cayNP{mikolov2013efficient}.}
    \label{SKIP}
\end{figure}

\begin{figure}[H]
    \centering
    \includegraphics[width=5cm]{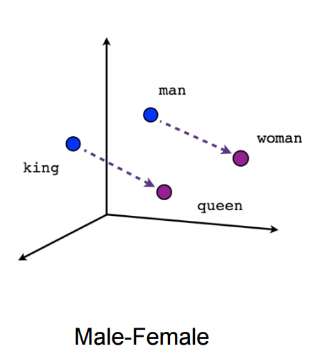}
    \caption{Representation of the relationship dynamics in space.}
    \label{REL}
\end{figure}

\subsection{Doc2Vec}
The Doc2Vec model is an extension of the Word2Vec model introduced by \cayNP{le2014distributed}. Indeed, the architecture is the same, but a new embedding is added, mapping a paragraph to a vectorial representation. This additional vector called “paragraph vector” represents the missing information from the current context. It acts like a memory of the topic of the paragraph. 

Similarly to the Word2Vec model, two representations are presented. The first one, called Paragraph Vector with Distributed memory is analogous to Continuous Bag of Words, and the second one called Paragraph Vector with Distributed Bag Of Words is analogous to Skip-Gram.

\begin{figure}[H]
    \centering
    \includegraphics[width=15cm]{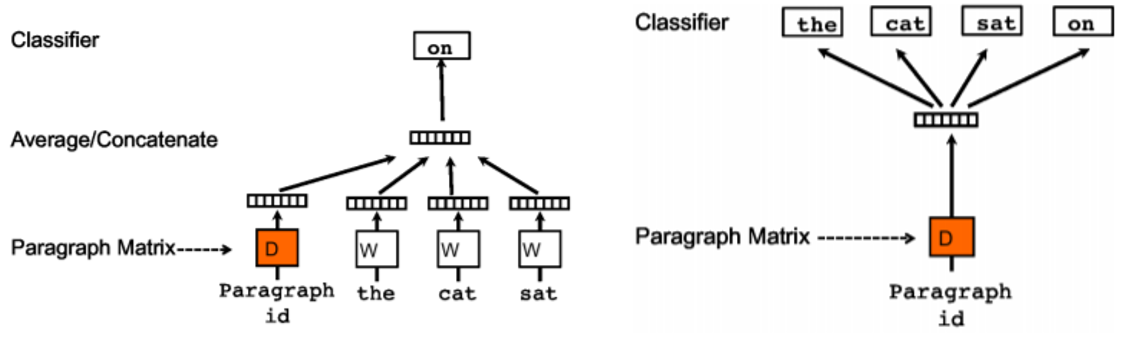}
    \caption{Illustration of the Paragraph Vector with Distributed Memory (on the left) and Paragraph Vector with Distributed Bag of Words (on the right) by \cayNP{le2014distributed}.}
    \label{PVDM}
\end{figure}

\subsection{From Attention to Transformers}

The comprehension of the following sections requires understanding of the following concepts, that won't be detailed in this paper.
\begin{itemize}
    \item Recurent Neural Networks (RNN): A type of neural networks that has access to past predictions when making a new prediction. The information about previous outputs is conveyed through "hidden states". RNN are able to work on sequential problems, which are problems that involve sequences of different sizes, such as translation. An article from \cite{RNN} explains the concept of RNN with further details, and a cheat sheet from \cite{CS230} summarizing the mathematical background around RNN is available on Stanford University's website. 
    \item Encoder-Decoder: Also called Sequence-to-Sequence (Seq2Seq), an architecture using RNNs introduced by \cayNP{sutskever2014sequence}. It performs well when the input's and output's lengths differ. An article from \cite{encoderdecoder} explains how the encoder-decoder architecture is built and how it allows better performances.
    
\end{itemize}

One of the main problems of using a Word2Vec representation is that the relations that are built between words are purely statistical, and reflect spatial proximity only. However, the purpose of NLP is to recreate subtle and complex relations that are put aside by statistical representations. Moreover, with the Recurrent Neural Networks 
that were heavily used lately for NLP tasks training can be very time-consuming. In order to  achieve efficiency, the computing processes involving training have to make the most of computer's parallel behavior (using GPUs), and compute the highest number of operations at the same time. The attention model introduced by \cayNP{vaswani2017attention}, aims to solve both of these problems by recreating the subtle links that exist within a sentence, and that are caught by the human brain when reading, and maximizing parallel computing. The general idea is to train as well a neural network and get from the hidden layers a transformation model that can be used for other tasks. 

\subsubsection{The Attention Mechanism}

Before the release of the \cayNP{vaswani2017attention} paper, a common approach to capture dynamic dependencies between words inside a sentence was to use Recurrent Neural Networks (RNNs) which are not detailed in this document. The idea of using such structures is to capture temporal dependencies between words. However, RNNs have some drawbacks. They allocates identical importance of words independently to the output to predict. In other words, the structure is not able to capture context and adapt the relation (weights) between words accordingly. \\

Attention models are neural networks layers that have been introduced to overcome that limitation. When looking at an input sequence, Attention is the mechanism that draws the relationships between the elements of the sequence. In other words, this mechanism emphasizes how every part of the input interact between themselves and how much influence one part has on another one. These interactions, which reproduce the context of the input, are represented in an Attention Matrix.\\ 

In other words, as mentionned by Mohammed Terry-Jack\footnote{\url{https://medium.com/@b.terryjack/deep-learning-the-transformer-9ae5e9c5a190}}, "an attention mechanism calculates the dynamic (alignment) weights representing the relative importance of the inputs in the sequence (the keys noted $K$) for that particular output (the query noted $Q$). Multiplying the dynamic weights (the alignment scores) with the input sequence (the values noted $V$) will then weight the sequence".

\begin{figure}[H]
    \centering
    \includegraphics[width=14cm]{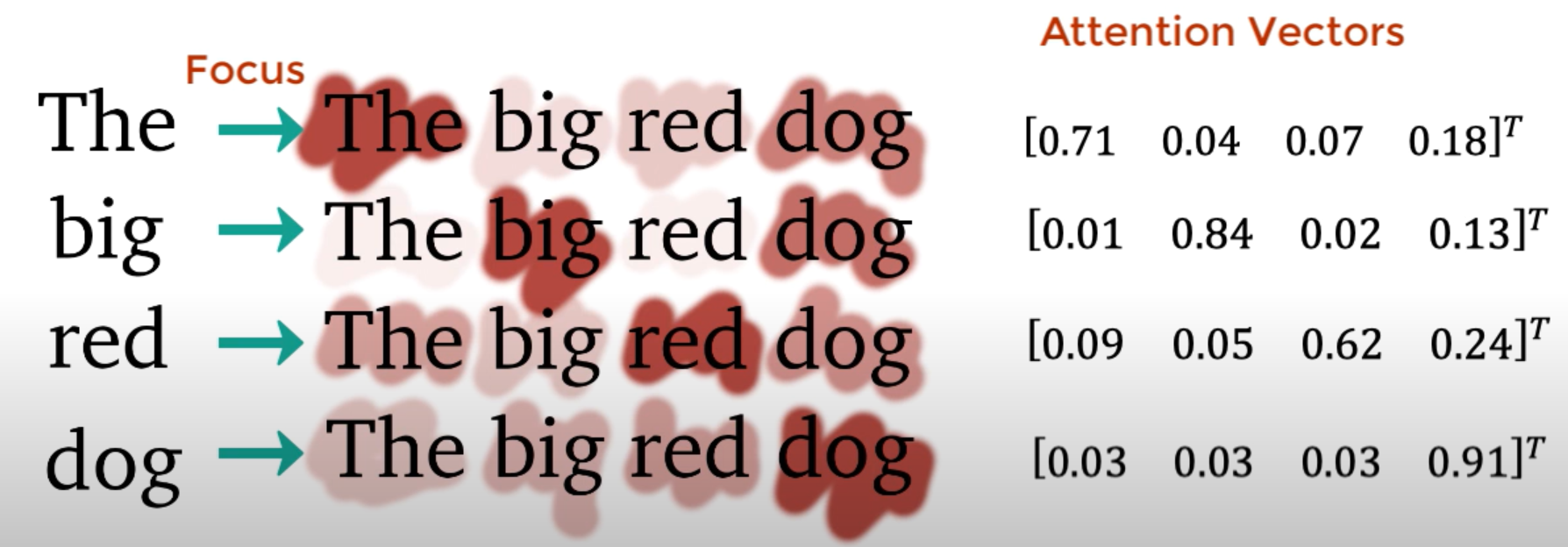}
    \caption{Illustration of the mechanism applied to a simple sentence by \cayNP{BERTajay}.}
    \label{attention}
\end{figure}

In the original paper \cay{vaswani2017attention}, two forms of Attention (so way to compute dynamics weights) are introduced. The first one is called "Scaled Dot-Product Attention" which is the application of matrix multiplications on context-based matrix. This operation outputs a new matrix giving information on inner interactions within the inputs. The second one called "Multi-Head Attention" is the concatenation of several Attention Matrices calculated on the same input, but with different parameters. This allows to have several representations of an Attention Matrix, which are then averaged, in order to reduce the risk of having a low-performance attention mechanism that would focus too much on the wrong parts of the input.

\begin{figure}[H]
    \centering
    \includegraphics[width=15cm]{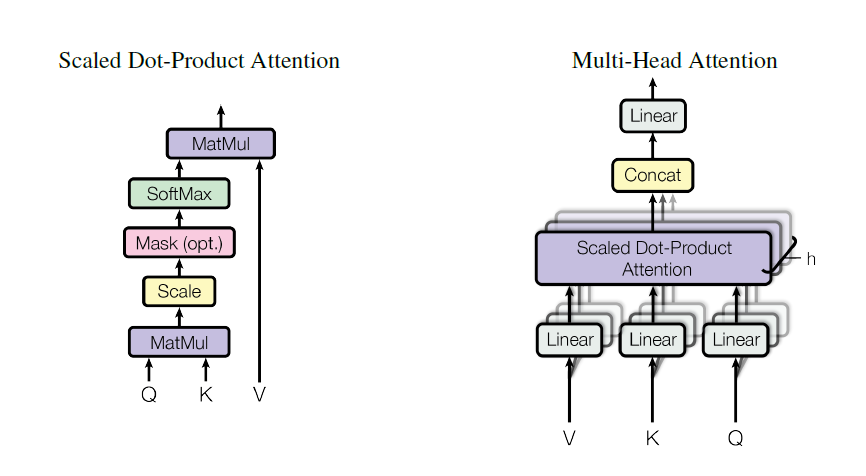}
    \caption{Architecture of the two attention mechanism as presented in the original paper\cay{vaswani2017attention}. Q, K and V are matrices that are components of the input.}
    \label{attention}
\end{figure}

Let's introduce some notations before going into further explanations. We will denote matrices dimension $\mathbb{R}^{n\times p}$ instead of $\boldsymbol{M}_{n\times p}(\mathbb{R})$. As the crux of the matter in this paper is Natural Language Processing, we will assume the inputs we are working with are textual. Let us denote a pre-processed input matrix $X\in\mathbb{R}^{n\times p}$ with $p\in\mathbb{N}$ the dimension of the embeddings, which is also the uniform size of all segment vectors (see definition of a segment section \ref{def} page \pageref{def}) and $n\in\mathbb{N}$ the length of the input sequence. $X$ has already gone through input embedding and positional embedding.

\subsubsection{Scaled Dot-Product Attention}
Before tackling Multi-Head Attention, let's see how the input matrix is being processed with Scaled Dot-Product Attention (number of heads $h=1$), by paying close attention to the dimensions of each matrix.

Let $W^Q\in\mathbb{R}^{p\times d_k} , W^K\in\mathbb{R}^{p\times d_k}, W^V\in\mathbb{R}^{p\times d_v}$, be three weight matrices from which we can obtain the following Q, K and V \footnote{Q,K and V stand for Query, Key and Value. They are parameters matrices that will be modified as the weight matrices will be adjusted during training. For more information about what each matrix represents, reader may refer to \cay{qkv}.} matrices :
\begin{center}
    $Q = X\cdot W^Q$      $\in\mathbb{R}^{n\times d_k}$
    
    $K = X\cdot W^K$      $\in\mathbb{R}^{n\times d_k}$
    
    $V = X\cdot W^V$      $\in\mathbb{R}^{n\times d_v}$
\end{center}
\begin{figure}[H]
    \centering
    \includegraphics[width=7.5cm]{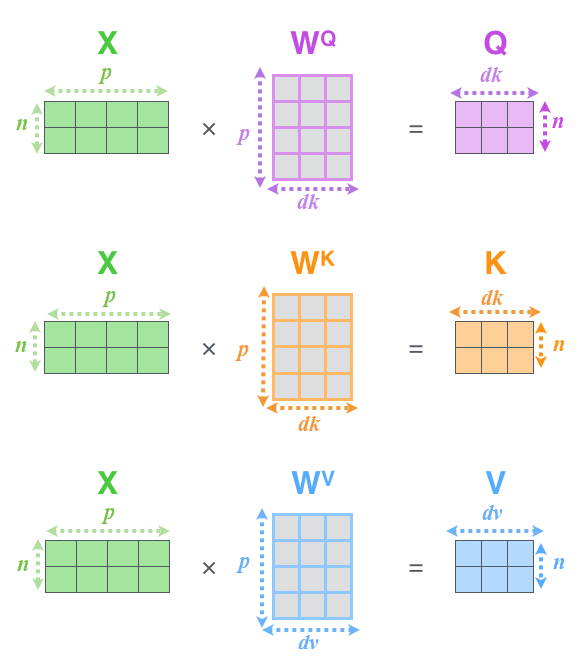}
    \caption{Illustration by \cayNP{JayAlammar}}
    \label{QKV2}
\end{figure}
Initially the values of the weight matrices are arbitrary, and will be adjusted later using a Feed-Forward Neural Network (see section \ref{FFN} page \pageref{FFN}). We now calculate the Attention using the formula from the original paper :

\begin{center}
    Attention$(Q,K,V)$ = softmax$(\frac{Q\cdot K^T}{\sqrt{d_k}})\cdot V$   $\in\mathbb{R}^{n\times d_v}$
\end{center}

Once the Attention matrix is calculated, it is multiplied by a matrix $W^0\in\mathbb{R}^{d_v\times p}$, so that the final output is a matrix :
\begin{center}
    $Z$   $\in\mathbb{R}^{n\times p}$
\end{center}

\subsubsection{Multi-Head Attention}
As it was briefly explained before, Multi-Head Attention is the concatenation of several Attention matrices calculated on the same input. Therefore the notations are the same as in the previous section, with an input matrix $X\in\mathbb{R}^{n\times p}$, but this time, the number of heads $h>1$. In the original paper, the actual value of $h$ is $8$.

In the previous part, only 3 weight matrices $W^Q, W^K, W^V$ were defined. However, Multi-Head Attention requires to define those three matrices for each head. Therefore, for $i\in\{0,..,h\}$ we will have $W^Q_i\in\mathbb{R}^{p\times d_k} , W^K_i\in\mathbb{R}^{p\times d_k}, W^V_i\in\mathbb{R}^{p\times d_v}$, with $d_k  = d_v = \frac{p}{h} $. 

We then calculate $Q_i\in\mathbb{R}^{n\times d_k}, K_i\in\mathbb{R}^{n\times d_k}, V_i\in\mathbb{R}^{n\times d_v}$ and $Z_i=$Attention$(Q_i, K_i, V_i)\in \mathbb{R}^{n\times d_v}$ for each head as done previously. 

Finally, the $Z_i$ are concatenated, giving a matrix Concat$(Z_i)\in \mathbb{R}^{n\times (d_v*h)}$, and the concatenation is multiplied by $W^0\in \mathbb{R}^{(d_v*h)\times p}$ in order to output the final matrix :

\begin{center}
    $Z$   $\in\mathbb{R}^{n\times p}$
\end{center}

\begin{figure}[H]
    \centering
    \includegraphics[width=15cm]{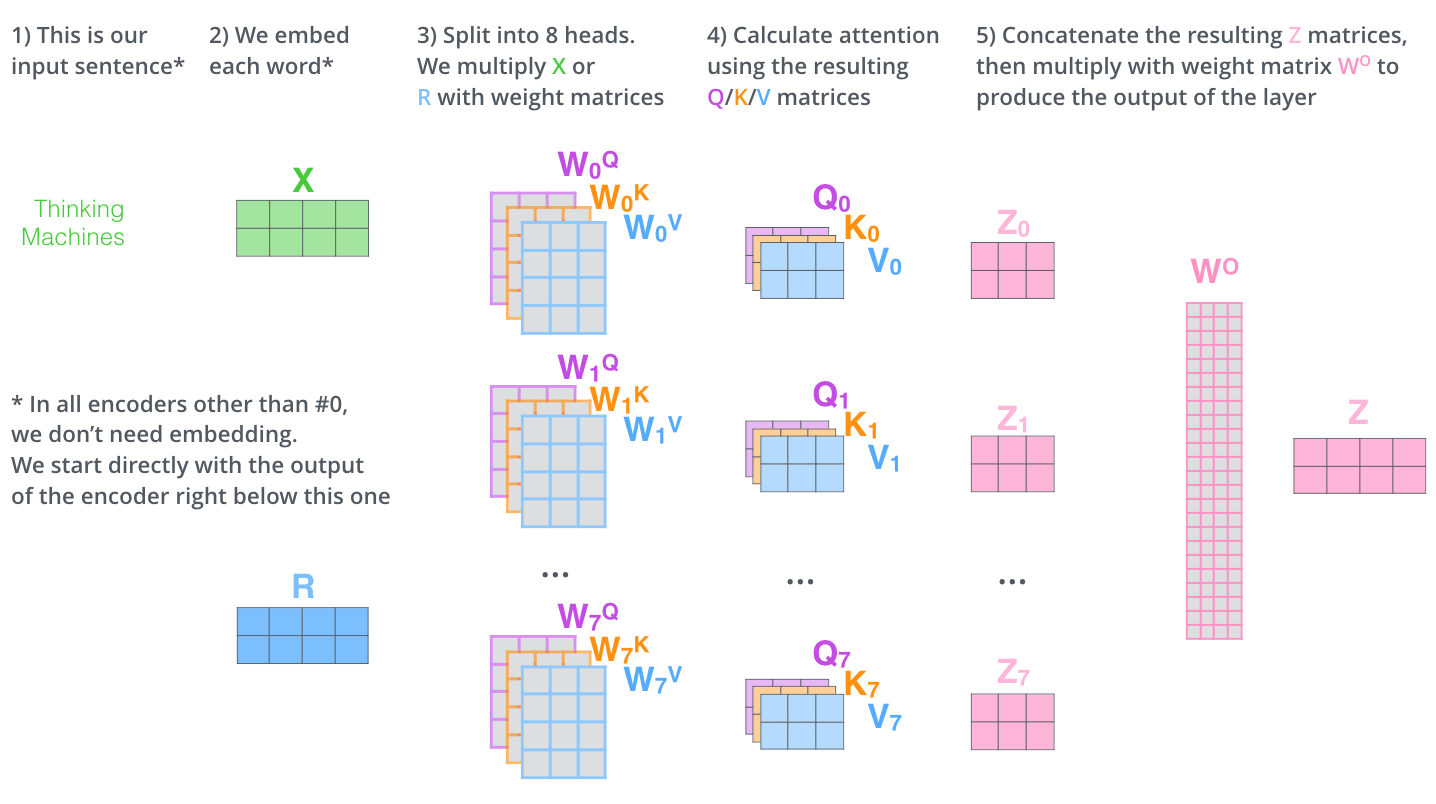}
    \caption{Step by step illustration of Multi-Head Attention by \cayNP{JayAlammar}.}
    \label{MHA}
\end{figure}

\subsubsection{Transformers Neural Network}
\label{transformerssection}

The Attention model was combined with neural networks in order to build a new type of model that essentially relies on attention: The Transformer Neural Network. This model is based on an encoder-decoder architecture.
Attention is used in the encoder part, the decoder part and the encoder-decoder part. This new type of model allows working on sequential problems without resorting to Recurrent Neural Networks (RNN), as Feed-Forward Neural Networks (FFN) are used at the top of both the encoder and the decoder. Using Feed-Forward Neural Networks instead of RNN improves calculation time and thus efficiency, allowing deeper and sturdier models to be trained using this technology.

\begin{figure}[H]
    \centering
    \includegraphics[width=7.5cm]{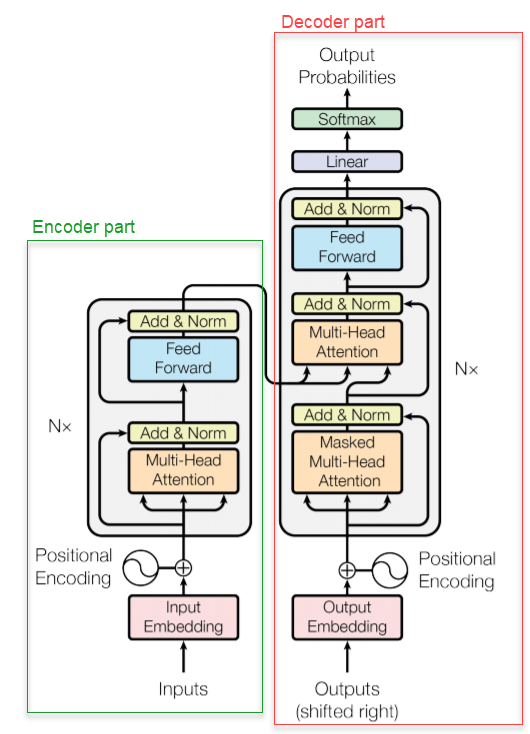}
    \caption{Architecture of the Transformer model from the original paper of \cayNP{vaswani2017attention}.}
    \label{Transformers}
\end{figure}

Before being fed to the Multi-Head Attention, the inputs go through two embeddings. The first one is a classic input embedding using one of the many methods available, like those listed in subsection \ref{TEsection} for text data. The second embedding is a Positional Encoding. This type of encoding aims to implement the notion of relative positions of the sequence (or time signal), like the order of words in a sentence, by indicating a numeric representation of a word's position in a sentence. Unlike RNNs, Feed-Forward Neural Networks won't have access to this information by their own, hence the need of positional encoding.

\subsubsection{Feed-Forward Neural Network (FFN)}
\label{FFN}
As shown on Figure \ref{Transformers}, the output of the Multi-Head Attention will be passed through a Feed-Forward neural network of the form : 
\begin{center}
    FFN$(x) = $max$(0, xW^1 + b_1)W^2 + b_2$
\end{center}

By back-propagation, training FFN$(C)$ will have the effect of adjusting the values of every $W^Q_i, W^K_i, W^V_i$ matrices needed for the calculation of $Z$, thus teaching attention to the model.

Eventually, the identified parameters for calibration are  $W^Q_i, W^K_i, W^V_i$. These parameters will be adjusted throughout the learning process of Transformers Neural Networks, and once it's done, we extract an operational attention model from the hidden layers. 

\subsection{Bidirectional Encoder Representation for Transformers (BERT)}
\label{bertsection}
BERT is one of the most recent models developed by \cayNP{devlin2018bert} for encoding inputs before addressing NLP tasks. The two main steps of training a BERT model are pre-training and fine-tuning, using Transformers Neural Networks. Pre-training is the part where the model learns about the language and how context operates in the said language, while fine-tuning addresses the specific issue to which the model is deployed. For comprehension purposes,  let us assume the task that is being studied is tweet sentiment analysis. 

In order to deal with theses two tasks, the BERT model's architecture is made of a series of encoders from the Transformers neural network architecture. In the previous paragraph, we identified $W^Q_i, W^K_i, W^V_i$ as the weight parameters to calibrate during training. As BERT's architecture is a succession of several encoders, there will be as many $W^Q_i, W^K_i, W^V_i$ matrices to calibrates as there are stacked encoders. Each of these matrices will be adjusted in a way that allows the model to understand the language it is trained on.

\begin{figure}[H]
    \centering
    \includegraphics[width=15cm]{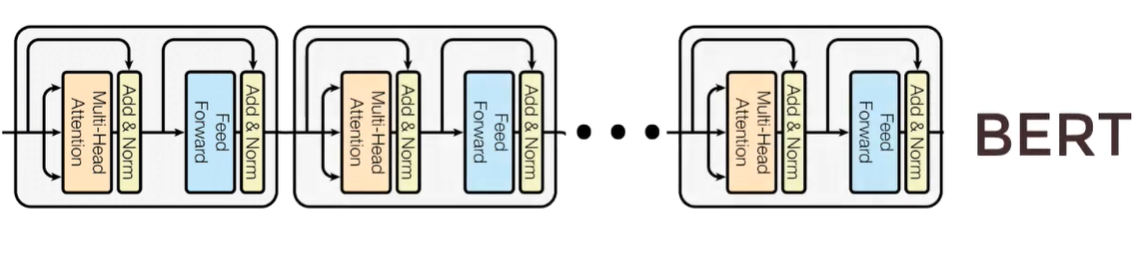}
    \caption{Stacking the encoder part from Figure \ref{Transformers} to build a BERT model. Illustration by \cayNP{BERTajay}.}
    \label{BERTTRANSFORMERS}
\end{figure}

\subsubsection{Inputs Representation}
Before going through pre-training and fine-tuning, let's see how the inputs are treated. In order to be processed by the model, the inputs are fed by pair of sentences $(A,B)$ in which the words are tokenized using a WordPiece embedding (see section \ref{WordPiece} page \pageref{WordPiece}). Moreover, each sequence starts with a special classification token $([CLS])$ and sentences are separated by a separation token $([SEP])$. 

In order to be processed by the model, a succession of embeddings is applied to the input (the tweets).
\begin{itemize}
    \item First, a special classification token ([CLS]) is added to the beginning of each sequence and sentences are separated with a separation token ([SEP]).	
    \item Then the inputs are tokenized using a WordPiece embedding (\cayNP{wu2016googles}), based on a vocabulary of 30,000 words. This type of tokenization splits a word int sub-words allowing words with similar sub-words to be attributed close semantic relations.
    \item A segment embedding is also applied to each token, denoting which sentence the word belongs to in the text.
    \item Finally, a positional embedding is applied, representing the location of the word in the text.
\end{itemize}
All of these operations combined provide the embedding applied to an input before being fed to the BERT model. 

\begin{figure}[H]
    \centering
    \includegraphics[width=15cm]{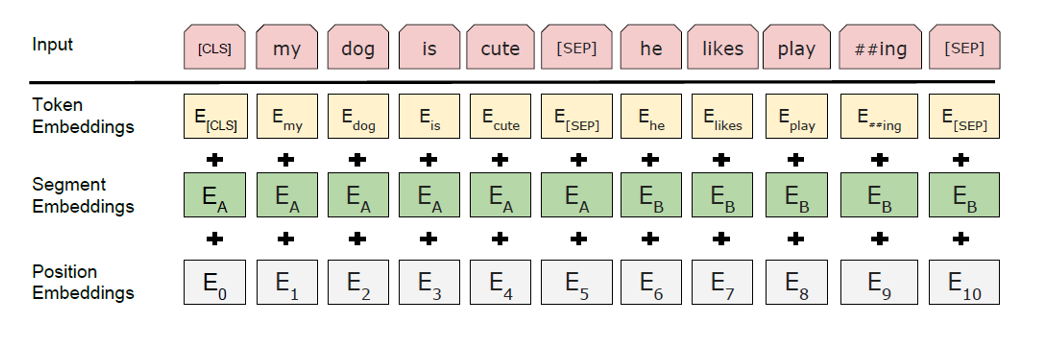}
    \caption{BERT input representation from the original paper \cay{devlin2018bert}.}
    \label{BERTINPUT}
\end{figure}

\subsubsection{Pre-training}

As the name itself suggests it, BERT is a combination of several encoders stacked together, extracted from Transformers neural networks.

This bidirectional encoder architecture allows the pre-training phase to take over two main tasks: Masked Language Modeling (MLM) and Next Sentence Prediction (NSP). During MLM, a sentence $d=(w_1, ...,w_n)$ is passed, where $w_i \in \mathbb{R}^{\mathbb{N}}, i\in \llbracket1,n\rrbracket$ are words. Approximately 15\% \footnote{The masking conditions are in reality a bit more complex than just hiding 15\% of the words. See the original article \cay{devlin2018bert} for more details.} of the words in each sequence are replaced with a $[MASK]$ token , and the model is trained to guess the masked tokens. The output vector $(O_1, ...,O_n)$ is then passed through a classification layer and eventually a softmax layer to obtain a distribution over the available vocabulary, indicating which word is more likely to be hidden behind the $[MASK]$ token.

\begin{figure}[H]
    \centering
    \includegraphics[width=10cm]{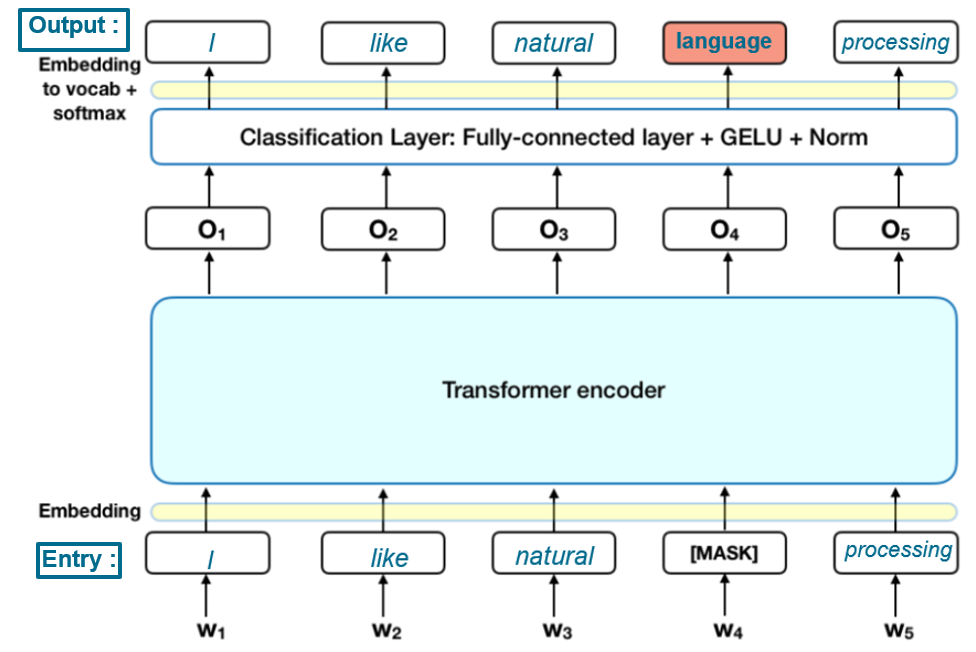}
    \caption{Masked Language Modeling with a GELU activation function illustrated by \cayNP{BERTtds}.}
    \label{MLM}
\end{figure}

The NSP part of pre-training is a binary classification problem. In this part, given two sentences $A$ and $B$, the model has to guess if sentence $B$ follows sentence $A$ or not. The output of the $[CLS]$ token will be passed through a binary classification layer and a softmax layer, similarly to MLM, and a vector $C$ of shape $2\times1$ will display the probability of the two sentences coming in succession.

\begin{figure}[H]
    \centering
    \includegraphics[width=7.5cm]{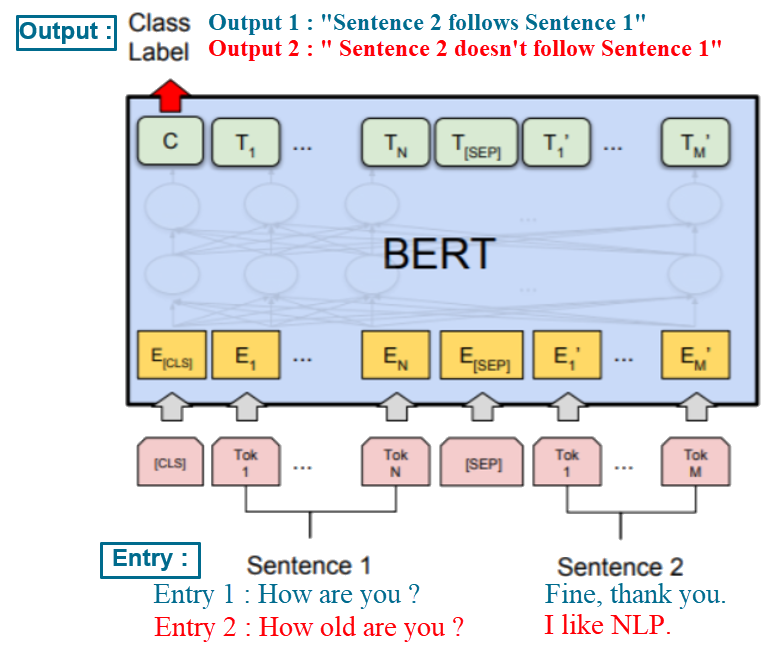}
    \caption{Next Sentence Prediction task illustrated by \cayNP{devlin2018bert}}
    \label{NSP}
\end{figure}

These two tasks are trained simultaneously, the aim of the pre-training phase being the minimization of the combined loss function of Masked Language Modeling and Next Sentence Prediction.

Text passages from BooksCorpus (800M words) \cay{zhu2015aligning} and English Wikipedia (2,500M words) are used to train the model on MLM and NSP.

\subsubsection{Fine-Tuning}

The fine-tuning phase on the other hand is quite easy to understand. It is the moment BERT uses the learnt language mechanics to solve a specific task. Just like in transfer learning, the input and output layer specific to our sentiment analysis task are added to the BERT model, and the parameters are adjusted accordingly. In our pre-training phase, the inputs are fed two by two  $(A,B)$. However, as the fine-tuning phase is task-specific, in our case the inputs will be a pair $(A,B)$ where $A$ = a tweet and $B = \emptyset$. For the outputs, a classification layer that will deliver the sentiment of the tweet is added at the end of the model. The $[CLS]$ representation will be fed to this new output layer and used for the sentiment prediction. 

\subsubsection{Review of the BERT model}

In the original paper, \cayNP{devlin2018bert} present two BERT models of different sizes : the base one, stacking 12 encoders from the Transformers representation, with a total of 110M parameters and the large one, stacking 24 encoders, with a total of 340M parameters. As the number of parameters of both models suggests, BERTs are quite brute force, but when compared to their competitors on tasks such as Sentence Classification or Question Answering, BERT models are the most efficient today. 

Although they achieve better results than other popular models on NLP tasks, they still take long to fine-tune, even when the uploaded models are already pre-trained. From it release in 2019, BERT has been extended and improved for many tasks. However, usages are limited by memory since new text representation models are more and more greedy.

\section{Reminder on machine learning metrics that can be used also for text mining}
\label{metrics}
Model evaluation metrics are required to quantify model performance.

\subsection{Classification Metrics}

To evaluate the performance of classification models, there are several possible metrics. Here, we define a few indicators in a non-exhaustive way.
Some common terms to be clear with are:
\begin{description}
    \item[$\bullet$ True positives (TP)] Predicted positive and are actually positive.
    \item[$\bullet$ False positives (FP)] Predicted positive and are actually negative.
    \item[$\bullet$ True negatives (TN)] Predicted negative and are actually negative.
    \item[$\bullet$ False negatives (FN)] Predicted negative and are actually positive.
\end{description}

\begingroup
\setlength{\tabcolsep}{6pt} 
\renewcommand{\arraystretch}{1} 
\begin{table}[H]
\begin{tabular}{|c|c|p{4.5cm}|p{3.5cm}|}
\hline
\multicolumn{1}{|c|}{Metrics} & \multicolumn{1}{c|}{Formula} & \multicolumn{1}{c|}{Meaning} & \multicolumn{1}{c|}{Range}\\\hline
Confusion matrix &
$\begin{bmatrix}
    TP & FN \\
    FP & TN
  \end{bmatrix}$
& A confusion matrix is an N$\times$N matrix, where N is the number of classes being predicted. It gives the counts of correct and incorrect classifications for each class. &  The higher the diagonal values of the confusion matrix the better. Conversely, the lower the values off the diagonal the better. All values are positive reals.\\\hline
Accuracy  & $\frac{TP + TN}{TP + TN + FP + FN}$ & Percentage of correctly classified instances out of the total predicted instances. & The accuracy is between 0 and 1, where 1 is the perfect score.\\\hline
Precision & $\frac{TP}{TP + FP}$ & Percentage of \textbf{positive instances} out of the total predicted positive instances. & As above.\\\hline
Recall    & $\frac{TP}{TP + FN}$ & Percentage of positive instances out of the \textbf{total actual positive instances}. &  As above.\\\hline
Specificity & $\frac{TN}{TN + FP}$ & Percentage of negative instances out of the \textbf{total actual negative instances}. & As above.\\\hline
F-score  & $\frac{(1+\beta^2)TP}{(1+\beta^2)TP+ \beta^2 FN + FP}$ & $\beta$ is a positive real, and it is chosen such that recall is considered $\beta$ times as important as precision. In practice $\beta$ is often set to 1, F1-score is the harmonic mean of precision and recall. & As above.\\\hline
ROC       &  & The Receiver Operator Characteristic curve represents the tradeoff between Recall and Specificity. & The closer a curve is to the top left corner the better.\\\hline
AUC       &  & AUC is the area under the ROC curve. & The AUC value lies between 0 and 1 where 1 indicates an excellent classifier and 0.5 the random model.\\\hline
\end{tabular}
\caption{Metrics for classification}
\end{table}
\endgroup

These metrics are valid for binary classification, for multi-label classification we can use micro, macro or samples averages.

The micro average allows to calculate the metric with TP, TN, FP, FN of the k-classes, for example for the precision :
$$PRE_{micro} = \frac{TP_1+...+TP_k}{TP_1+...+TP_k+FP_1+...+FP_k}$$
The macro average is obtained by averaging the metric of each class: 
$$PRE_{macro} = \frac{PRE_1+...+PRE_k}{k}$$
The samples average returns the average metric of each instance.

All these metrics need to be compared with a baseline model. A great score doesn't mean that the model performs well if a random model has also a good score. A straightforward baseline model is the dummy classifier that always predict the same result.  

\subsection{Regression Metrics}

The regression task, unlike the classification task, outputs continuous value within a given range.

\begingroup
\setlength{\tabcolsep}{6pt} 
\renewcommand{\arraystretch}{2} 
\begin{table}[h!]
\begin{tabular}{|c|c|p{7cm}|p{2cm}|}
\hline
\multicolumn{1}{|c|}{Metrics} & \multicolumn{1}{c|}{Formula} & \multicolumn{1}{c|}{Meaning} & \multicolumn{1}{c|}{Range}\\\hline
MAE & $\frac{1}{N}\sum_{i=1}^N|y_i - \hat y_i|$
& The Mean Absolute Error is the average of the difference between the actual values and the predicted values. & $[0; +\infty[$ \\\hline
MSE & $\frac{1}{N}\sum_{i=1}^N(y_i - \hat y_i)^2$
& The Mean Squared Error measures the square of the difference between the actual values and the predicted values. & $[0; +\infty[$ \\\hline
RMSE & $\sqrt\frac{\sum_{i=1}^N(y_i - \hat y_i)^2}{N}$
& The Root Mean Squared Error measures the average magnitude of the error by taking the square root of the average of squared differences between prediction and actual observation. & $[0; +\infty[$ \\\hline
$R^2$ & $1-\frac{MSE(model)}{MSE(baseline)}$
& The Coefficient of Determination or $R^2$ compares the current model with a constant baseline. &  $]-\infty; 1]$ \\\hline
\end{tabular}
\caption{Metrics for Regression}
\end{table}
\endgroup

\subsection{Business Metrics}

In addition to statistical metrics, model fitting and selection must be completed by business assessment. It is hard to define exhaustive list of business metrics but as an example, using NLP for claims adjudications has also been assessed in term of losses for claims paid that should not have been versus, claims we can decline more efficiently thanks to scoring. Projecting algorithms decision into business metrics (like estimated loss ratio or A/E ratio in insurance) is key. It is worth to inform reader about it since to bring value, text mining algorithms should not only be integrated in technological solutions but need also to be calibrated to optimize an operational metric.

\section*{Conclusion}
This survey presented different processes and techniques that are common in Natural Language Processing. Research in this area has been tremendous since the past ten years. BERT seems today to delimit a new era in the usage of textual information. The frequency of new models releases lead to continuously adapt text mining algorithms but also provide a large amount of new opportunities in the insurance industry.

\clearpage

\end{document}